\documentclass[twocolumn]{article}
\usepackage[utf8]{inputenc}
\usepackage{authblk}       % For author and affiliation formatting
\usepackage{geometry}      % Page margins
\usepackage{cite}          % Clean citation styles

\usepackage{hyperref}       % hyperlinks
\usepackage{url}            % simple URL typesetting
\usepackage{booktabs}       % professional-quality tables
\usepackage{nicefrac}       % compact symbols for 1/2, etc.
\usepackage{microtype}      % microtypography
\usepackage{graphicx}
\usepackage{amsmath}
\usepackage{float}
\usepackage{subcaption}
\usepackage{amssymb}
\usepackage{breqn}
\usepackage{multirow}
\usepackage{xspace}
\usepackage{mathtools}
\usepackage{adjustbox}
\usepackage{color}
\definecolor{transparent_highlight}{RGB}{255, 255, 0}
\usepackage{framed,multirow}
\usepackage{amssymb}
\usepackage{latexsym}
\usepackage{url}
\usepackage{xcolor}
\usepackage{hyperref}
\usepackage{soul, color,colortbl} % for table backgrounds
\definecolor{light-gray}{gray}{0.70}
\definecolor{newcolor}{rgb}{.8,.349,.1}
\definecolor{Gray}{gray}{0.9} % for table backgrounds

\newcommand{\review}[1]{{#1}} % remove highlight for review

\newcommand{\LDSC}{\mathcal{L}_{\mathrm{DSC}}}
\newcommand{\LCE}{\mathcal{L}_{\mathrm{CE}}}

\newcommand{\Lsimi}{\mathcal{L}_{\mathrm{similarity}}}
\newcommand{\LgReg}{\mathcal{L}_{\mathrm{gReg}}}
\newcommand{\Lsmo}{\mathcal{L}_{\mathrm{smooth}}}
\newcommand{\Lcyc}{\mathcal{L}_{\mathrm{cyclic}}}
\DeclareRobustCommand{\textsupsub}[2]{{%
  \m@th\ensuremath{%
    ^{\mbox{\fontsize\sf@size\z@#1}}%
    _{\mbox{\fontsize\sf@size\z@#2}}%
  }%
}}

\hypersetup{
    colorlinks,
    linkcolor=blue,
    citecolor=blue,
    urlcolor=blue,
    filecolor=blue
}
\title{\textbf{CMRINet: Joint Groupwise Registration and Segmentation for Cardiac Function Quantification from Cine-MRI}}
\author[1]{Mohamed S. Elmahdy}
\author[1]{Marius Staring}
\author[1]{Patrick J. H. de Koning}
\author[2,3]{Samer Alabed}
\author[2]{Mahan Salehi}
\author[2]{Faisal Alandejani}
\author[2]{Michael Sharkey}
\author[4]{Ziad Aldabbagh}
\author[2,3]{Andrew J. Swift}
\author[1]{Rob J. van der Geest\thanks{Corresponding author: Division of Image Processing, Department of Radiology, Leiden University Medical Center, The Netherlands. Email: \texttt{R.J.van\_der\_Geest@lumc.nl}}}
\affil[1]{Division of Image Processing, Department of Radiology, Leiden University Medical Center, The Netherlands.}
\affil[2]{Department of Infection, Immunity and Cardiovascular Disease, University of Sheffield, Sheffield, UK.}
\affil[3]{INSIGNEO, Institute for In Silico Medicine, University of Sheffield, Sheffield, UK.}
\affil[4]{Department of Clinical Radiology, Sheffield Teaching Hospitals, Sheffield, UK.}
\date{}

\begin{document}
\twocolumn[
  \maketitle
  \begin{center}
    \begin{minipage}{0.8\textwidth}
      \begin{abstract}
        Accurate and efficient quantification of cardiac function is essential for the estimation of prognosis of cardiovascular diseases (CVDs). One of the most commonly used metrics for evaluating cardiac pumping performance is left ventricular ejection fraction (LVEF). However, LVEF can be affected by factors such as inter-observer variability and varying pre-load and after-load conditions, which can reduce its reproducibility. Additionally, cardiac dysfunction may not always manifest as alterations in LVEF, such as in heart failure and cardiotoxicity diseases. An alternative measure that can provide a relatively load-independent quantitative assessment of myocardial contractility is myocardial strain and strain rate. By using LVEF in combination with myocardial strain, it is possible to obtain \review{a thorough description of cardiac function}. Automated estimation of LVEF and other volumetric measures from cine-MRI sequences can be achieved through segmentation models, while strain calculation requires the estimation of tissue displacement between sequential frames, which can be accomplished using registration models. These tasks are often performed separately\review{, potentially limiting the assessment of cardiac function.} To address this issue, in this study we propose an end-to-end deep learning (DL) model that jointly estimates groupwise (GW) registration and segmentation for cardiac cine-MRI images. The proposed anatomically-guided Deep GW network was trained and validated on a large dataset of 4-chamber view cine-MRI image series of 374 subjects. A quantitative comparison with conventional GW registration using \texttt{elastix} and two DL-based methods showed that the proposed model improved performance and substantially reduced computation time.

        \vspace{0.5em}
        \vspace{2em}
      \end{abstract}
    \end{minipage}
  \end{center}
]

\section{Introduction}
Cardiovascular disease (CVD) is the leading cause of mortality \cite{virani2020heart} worldwide, with approximately 17.9 million deaths attributed to CVD in 2019, according to the World Health Organization. A significant proportion of these deaths occur in low- and middle-income countries. Therefore, accurate and efficient quantification of cardiac function is critical for early prognosis and treatment as well as  preventing the development of irreversible complications. Cardiac function integrity encompasses both normal morphology as well as ordinary mechanical characteristics represented by the pumping function of the heart's different chambers. One of the most widely used indices for assessing cardiac pumping function is left ventricular ejection fraction (LVEF). However, the estimation of LVEF is subject to influence from factors such as inter-observer variability and varying  pre-load and after-load conditions, which can reduce its reproducibility \cite{marwick2018ejection}. Additionally, cardiac mechanical dysfunction may not always manifest as alterations in LVEF, as seen in conditions such as heart failure with preserved ejection fraction (HFpEF) \cite{obokata2020diastolic}, cardiotoxicity \cite{ewer2008left}, cardiomyopathies \cite{saito2012clinical, hsiao2013speckle, phelan2012relative}, and congenital heart disease \cite{diller2012left}. Alternatively, myocardial strain and strain rate can provide a relatively load-independent quantitative assessment of myocardial contractile function.  
Among the various imaging modalities, tagged-MRI is considered the reference standard for strain quantification \cite{kim2004myocardial, moore2000quantitative, woo2014tissue}. The main advantage of tagging is that deformation is directly measured by physical properties of the tissue. However, tagged-MR is limited by factors such as tag-line degradation and time-consuming image post-processing and the fact that it requires an additional dedicated acquisition sequence. Therefore, its usage has been limited to research purposes and the availability of commercial tagging analysis software solutions has been limited as well \cite{amzulescu2019myocardial}. Alternatively, myocardial strain can be estimated from echocardiography and cine-MRI using feature tracking techniques \cite{claus2015tissue, pedrizzetti2016principles}. In this paper we focus on strain estimation from cine-MRI, as it does not require dedicated sequences and is ubiquitously acquired in daily clinical practice. 

The accurate quantification of cardiac function relies on the successful completion of both cardiac segmentation and motion estimation tasks. These tasks are critical for obtaining regional measures such as strain, as well as global measures like LVEF. Historically, these tasks were performed separately in clinical workflows. However, they are closely related and complement each other. By combining them in the workflow, the quantitative assessment of cardiac function can be made more efficient and effective \cite{cheng2017segflow, elmahdy2021joint}. 

In a real clinical workflow scenario, only the end-diastolic and end-systolic frames of a cine-MRI sequence are typically manually annotated. However, motion estimation of the walls of different heart chambers requires consistent and robust delineation of these regions across all frames, which can be challenging \cite{claus2015tissue}. Manual delineation of these regions is time-consuming and often results in intra- and inter-observer variability \cite{risum2012variability, schuster2015cardiovascular} and it requires highly trained personnel which could be expensive. Automatic delineation methods, such as segmentation approaches or contour propagation algorithms \cite{wang2019fully, galati2022accuracy} can alleviate these issues. However, segmentation based automatic methods do not consider underlying mechanical properties of the cardiac tissue such as incompressibility, elasticity and periodicity, which is important for accurate and robust strain estimation \cite{de2012temporal, mansi2011ilogdemons}. On the other hand, registration methods can model these biomechanical properties and constrain the solution accordingly. Therefore, combining segmentation and registration tasks can be beneficial for accurate and robust strain estimation. 

Motion estimation from a time-series of images can be formulated as a pairwise (PW) or a groupwise (GW) registration problem. With PW registration, images are registered sequentially either against the next image in the sequence or against a selected reference image, with the other images registered in pairs against that reference. In contrast, GW registration optimizes a set of images concurrently to a common hidden space using a single optimization procedure, taking into account information from all the images in the sequence. The reference in this case is created using information from the entire temporal sequence. GW registration yields better results than PW, as the optimization is solved as a whole. However, this comes at a high computational cost due to the iterative nature of classical GW optimization.

Segmentation and registration are correlated tasks, therefore learning a meaningful representation from one task could be helpful for the other task. In addition, modeling these correlated tasks in a single end-to-end model introduces inductive bias and helps regularize each task. Notably, learnable groupwise registration has its limitations; for instance, outliers or subjects with abnormal motion patterns can result in inaccurate or suboptimal registration outcomes.

To address these limitations, jointly modeling segmentation and groupwise registration tasks for cardiac cine-MRI could have several advantages over individual GW registration or segmentation networks. One key benefit is that the joint approach can improve the accuracy of image alignment by taking into account the spatial, temporal, and anatomical relationships between images. This is particularly useful when there is significant motion or variability across the dataset. Another advantage is that jointly segmenting and registering all images at once can be more computationally efficient compared to performing each step separately. Additionally, joint approaches can also improve the consistency and reproducibility of segmentation results by reducing variability and errors caused by individual segmentation. Moreover, joint segmentation and groupwise registration approaches can increase the statistical robustness of the results in situations where the number of subjects is limited. Overall, joint segmentation and groupwise registration approaches for cardiac cine-MRI offer a powerful tool for improving the accuracy and consistency of image alignment and segmentation.

In recent years, deep learning-based models have become widely recognized as powerful tools in the field of cardiac MRI \cite{leiner2019machine, hernandez2021deep, chen2020deep}. For segmentation, \cite{bai2018automated} trained a 2D fully connected network (FCN) on a large dataset ($\sim$5000 subjects) for short-axis (SAX), 2-chamber and 4-chamber view segmentation. \cite{tao2019deep} proposed a multi-vendor, multi-center CNN for LV endocardial and epicardial segmentation in SAX cine-MRI. \cite{khened2019fully} developed a multi-scale residual denseNet for LV and RV segmentation from the SAX view. To exploit the temporal characteristic of cine-MRI, \cite{poudel2016recurrent} proposed a 2D FCN with a recurrent neural network (RNN) to model the inter-slice coherence in the SAX view. \cite{wang2021dense} also proposed an RNN architecture, but for the LAX view. Recently, \cite{suinesiaputra2021deep} explored the effectiveness of U-Net for segmentation and landmark detection for legacy datasets including SAX and LAX views. \review{\mbox{\cite{ruijsink2020fully, pierpaolo2020left}} proposed a CNN segmentation network for SAX view segmentation and performed a volumetric analysis as well as strain analysis. However, the strain analysis was a separate function and derived from the segmentation results.} Most of the literature focuses on LV segmentation from the SAX view \cite{chen2020deep}, while only few studies have been reported on 4-chamber LAX view segmentation. This may be due to the availability of public datasets on SAX views such as the Sunnybrook Cardiac Dataset \cite{Radau2009}. 

For motion estimation via registration, \cite{zhang2021groupregnet} proposed GroupRegNet, a one-shot GW registration network that estimates respiratory motion from thorax 4D-CT. \cite{qiao2020temporally} proposed motionNet, a CNN network that tracks the myocardial displacement between consecutive frames. Using motionNet as a building block, the authors introduced a network called groupwise MotionNet. \cite{de2019deep} proposed a multi-scale PW affine and deformable registration network. \cite{krebs2019learning} proposed a probabilistic model for diffeomorphic PW registration of SAX cine-MRI. To the best of our knowledge, no work on end-to-end DL-based GW registration has been proposed for application in cine-MRI data. 

Finally, for joint segmentation and motion estimation, \cite{qin2018joint} proposed to jointly optimize a PW registration and a segmentation network by sharing the weights of the network encoder as well as the loss function for SAX cine-MRI. In earlier work, we proposed a joint segmentation and PW registration network through a cross-stitch architecture \cite{elmahdy2021joint} for evaluation of multi-time point imaging data in prostate cancer patients. Recently, \cite{morales2021deepstrain} proposed a deep learning workflow called DeepStrain for strain estimation from SAX cine-MRI. In that study, the authors proposed three DL networks for localization, segmentation, and PW registration. However, these three networks were trained separately and not jointly optimized, which does not guarantee consistence between the segmentation and registration results. 

\review{From the aforementioned literature review it is clear that automated assessment of cardiac function has been a subject of extensive research, primarily focusing on methodologies applied to the short-axis (SAX) view. However, there remains a dearth of comprehensive analyses specifically targeting the less-explored 4-chamber LAX view. This view provides a distinct perspective of cardiac function, yet existing methodologies predominantly concentrate on SAX views, leaving a gap in the comprehensive assessment of cardiac function. Current approaches often rely on separate segmentation and analysis steps or non-deep learning-based methods for strain estimation after segmentation, resulting in segmented contours' dependency and limited simultaneous estimation of local and global metrics.}

In this study, our aim is to address this gap by developing an end-to-end network that can perform segmentation and groupwise registration of cardiac 4-chamber LAX view cine-MRI. The proposed architecture consists of two networks: SegNet and Group-RegNet, which are jointly optimized for cine-MRI segmentation and alignment. The network is trained and validated on a large cohort of subjects with a wide range of pathologies. The network is fully automatic and does not require the availability of any annotated reference frame. We optimized the SegNet and Group-RegNet networks using different loss functions, training schemes, and implicit template estimation approaches. We also introduced a dictionary of masks for the cine-MRI. Additionally, we clinically validated the proposed method by estimating global longitudinal strain and LVEF. \review{To the best of our knowledge, this is among the initial end-to-end networks that jointly optimize segmentation and motion estimation for the automated quantification of cardiac function, integrating both tasks within a unified framework.}

 The remainder of this paper is structured as follows: Section \ref{methods_Section} presents the base network architecture, segmentation network, groupwise registration network and the joint approach for segmentation and groupwise registration. In Section \ref{materials_section}, we describe the dataset and details about the implementation as well as the evaluation metrics. In Sections \ref{results_section} and \ref{discussion_section}, we present our findings and discuss the results. Our conclusion is summarized in Section \ref{conclusion_section} along with potential directions for future research.
\section{Methods} \label{methods_Section}

\begin{figure*}[t]
\begin{center}
\includegraphics[width=\linewidth]{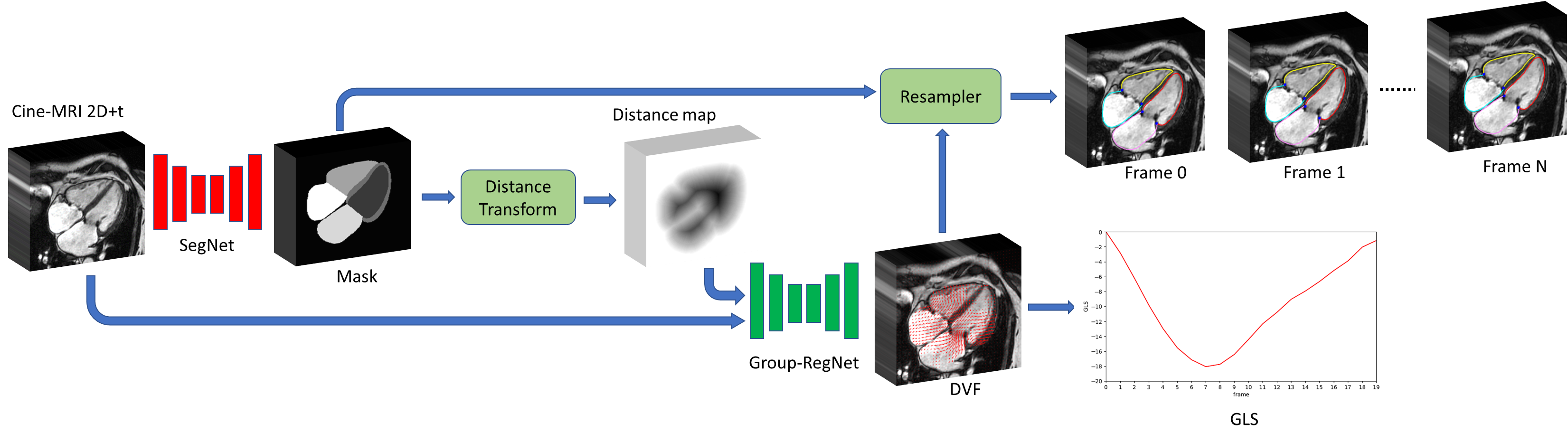}
\caption{The workflow for the proposed CMRINet network.}
\label{fig:workflow}
\end{center}
\end{figure*}

In this paper we propose an end-to-end joint segmentation and GW registration network enabling fully automated quantification of cardiac function. In the following section, we introduce the base network architecture as well as the segmentation and GW registration networks that together comprise our proposed network as shown in Fig. \ref{fig:workflow}. 

\subsection{Base Network Architecture}
The base architecture for the networks presented in this paper is a 2D residual U-Net, which is inspired by the work of \cite{zhang2018road}. The Res-UNet combines the strengths of both the U-Net \cite{ronneberger2015u} and the residual network \cite{he2016deep}. The Res-UNet has three main components: encoding, skip connections and decoding. Through the encoding part, the input is encoded into dense representations, while the decoding part restores the depictions to the final prediction. Skip connections link the encoding and decoding branches to preserve the high resolution features. The network encodes the input through a series of residual blocks. Each block of the encoding and decoding parts are comprised of a residual block. Each residual block consists of one convolutional block and an identity mapping, which connects the input and the output of the block. The convolutional block has a convolution layer of size $3\times3$, instance normalization, and a rectified linear unit (ReLU). We used bilinear interpolation for both down-sampling and up-sampling in the encoding and decoding branches, respectively. The network has four convolutional blocks in both the encoder and decoder, where the number of filters starts from 32 and doubles at each resolution. 

\subsection{Segmentation Network} \label{segNetwork}
The input to the segmentation network is a series of cine-MRI frames stacked in a single batch. 
We experimented training the network using the Dice Similarity Coefficient (DSC) as well as cross-entropy (CE) loss functions. DSC loss quantifies the overlap between the network prediction S$^{\mathrm{pred}}$ and the groundtruth S$^{\mathrm{gt}}$ as follows:
\begin{align}
\centering
\LDSC = 1 - \frac{1}{K}\sum^{K}_{k=1}
        \frac{2*\sum_{\Omega}S^{\mathrm{pred}}_k(\Omega) \cdot S^{\mathrm{gt}}_k(\Omega)}
        {\sum_{\Omega}S^{\mathrm{pred}}_k(\Omega)+\sum_{\Omega}S^{\mathrm{gt}}_k(\Omega)},
\label{eq:DiceLoss}
\end{align}
where $\Omega$ is the set of pixel coordinates and $K$ is the number of structures to be segmented (see Section \ref{dataset}). On the other hand, the CE loss measures the difference between the predicted probability distribution, $p_i$, and the true probability distribution of the output, $y_i$:
\begin{align}
\centering
\LCE = -\sum_{i=1}^{K} y_{i} \log(p_{i}).
\label{eq:CELoss}
\end{align}

The proposed segmentation network is dubbed SegNet and has 1,882,150 trainable parameters. 

% , $S^{\mathrm{gt}}_k$ is the ground truth segmentation, and $S^{\mathrm{pred}}_k$ the predicted probabilities. 

\subsection{Groupwise Registration Network} \label{groupRegNetwork}
Groupwise registration (GW) refers to the simultaneous alignment of a set of images to a shared space using a single optimization process, with consideration given to information from all of the images. Let $I_N$ denote a group of images $I_N = \{ I_n|n=1, \ldots, N\}$, where $I_n$ represents each image in the group. In this paper, each 2D image represents one time point across the cardiac cycle from a $2D+t$ cine-MRI sequence. The objective of the GW registration network is to predict a set of \review{displacement vector fields ($\phi_N$)}. The optimization function for GW registration can thereby formulated as:
\begin{align}\label{eq:groupReg}
\centering
% \argmin_{T^N_{temp}} \: \{ \: \Lsimi + \lambda_0 \Lsmo +
% \lambda_1 \Lcyc\:\},
% \argmin \: \{ \: \Lsimi + \lambda_0 \Lsmo +
% \lambda_1 \Lcyc\:\},
\LgReg &= \Lsimi(T_N\circ I_N, I_{temp}) \\ 
&+ \lambda_0 \: \Lsmo(T_N) + \lambda_1 \: \Lcyc(T_N), \notag \\
T_N(x) &= \phi_N(x) + x,  
\end{align} 

\noindent where $T_N$ is a set of transformations $\{ T_n|n=1, \ldots, N\}$ that align the coordinates in the estimated template to the corresponding coordinates in the input images. $T_N\circ I_N$ refers to the warped input image, where $I_{temp}$ is the implicit template estimated by averaging all the warped input images. $\lambda_0$ and $\lambda_1$ are the weights for the smoothness and cyclic regularization losses. The similarity loss used in this paper to measure the similarity between the template image and the warped input images is the local normalized cross correlation (LNCC) as shown in Eq. (\ref{eq:similarity}). \review{We chose LNCC instead of NCC or MSE losses due to its robustness against local intensity shifts and noise. Moreover, it can be implemented efficiently using the convolution operation.} 
\begin{align}
\centering
\Lsimi &= 1 -  \frac{1}{N}\sum_n \mathrm{LNCC}\left(T_N\circ I_N, I_{temp}\right),\label{eq:similarity} \\
I_{temp} &= \frac{1}{N} \sum_{n=1}^{N} T_n\circ I_n \label{eq:template}
\end{align} 
 
In order to encourage the network to predict a realistic and smooth displacement, a smoothness regularization loss is introduced. This loss penalizes the $\ell_1$ norm of the temporal total variation of the displacement as follows:
\begin{equation}\label{eq:smooth}
\centering
\Lsmo = \frac{1}{2N|\Omega|}\sum_{n,x \in \Omega, i \in X, Y} \| \nabla_i  \phi_n(x)\|_1,
\end{equation}

\noindent \review{where $\nabla_i  D_n(x)$ is the partial derivative of the displacement field with respect to axis $i$.} Since the cine-MRI sequences in this study encompass a complete cardiac cycle, it is valid to assume periodicity in the estimated displacement. This can be imposed by the cyclic constraint. This term constrains the sum of the displacements across the cine-MRI sequence to be zero:
\begin{equation}\label{eq:cyclic}
\centering
\Lcyc = \sqrt{\frac{1}{2N|\Omega|}\sum_{x \in \Omega}  \left( \sum_n T_n(x)\right)^2}.
\end{equation}

The network estimates $T_N(x)$ that align the input images to the template. However, in order to estimate the displacement $T^{i}_{j}$ that maps any $i_{th}$ image to $j_{th}$ image requires the inverse transformation $T^{-1}_{N}(x)$. This inverse transformation is estimated numerically using the fixed-point method \cite{chen2008simple}. Now $T^{i}_{j}(x)$ can be calculated using the composition of the forward and inverse transformations as follows:
\begin{equation}
 T^{i}_{j}(x) = T_i(T^{-1}_{j}(x))
\end{equation}

The input to the network is a cine-MRI sequence stacked in the channel dimension. Our motion estimation network is a 2D network with 1,890,642 trainable parameters and referred to as Group-RegNet throughout the rest of the paper. 

\subsection{Proposed CMRINet}
In previously introduced joint networks \cite{elmahdy2021joint, mahapatra2018joint, JrsGan}, segmentation and registration tasks were optimized in a parallel manner, hence the manual annotations were only incorporated into the training process as binary masks via the loss function. However, one of the challenges in intensity-based cine-MRI registration is that the network can become distracted by the blood flow related intensity fluctuations during the cardiac cycle, rather than focusing on the deformation of the walls of the cardiac chambers, particularly in the region near the mitral and tricuspid valves. To address this issue, we propose to train the segmentation and GW registration in a sequential manner. This enables the integration of the generated contours into the registration in the form of distance maps, rather than via binary masks. Registering the contours' distance maps alongside the cine-MRI images can potentially be more robust to variations in image intensity, noise, contrast, resolution and artifacts, which are very common in the dataset as shown later in Section \ref{dataset}. Hence, the network learns to focus on the intensity variations as well as the walls of the cardiac chambers.  

Here, we propose an end-to-end fully automated joint segmentation and GW registration network dubbed CMRINet. The structure of the proposed network is shown in Figure \ref{fig:workflow}. CMRINet is comprised of two networks: SegNet and Group-RegNet, which are jointly optimized to segment and register a cine-MRI sequence. Although SegNet is a fully automatic segmentation network, it can only be used for quantifying volumetric parameters such as end-systolic (ES) volume, end-diastolic (ED) volume and ejection fraction. On the other hand, Group-RegNet can only quantify the cardiac deformation and thus strain and is considered semi-automatic because it requires the availability of the manual annotation of either the ES or ED frames. Hence, both networks can only partially quantify cardiac function. 
 
In contrary to traditional semi-automatic GW registration networks that require the availability of manual annotation for one of the sequence frames, CMRINet is fully automated. First, the cine-MR sequence is segmented using SegNet and subsequently the Euclidean distance maps of the predicted segmentations are computed. The distance maps alongside the cine-MRI sequence are then registered using Group-RegNet network. The predicted displacement vector fields (DVF) are used to propagate the segmentation of the ED phase estimated by SegNet to all other phases. 

\begin{figure}[t]
\begin{center}
\includegraphics[width=1\linewidth]{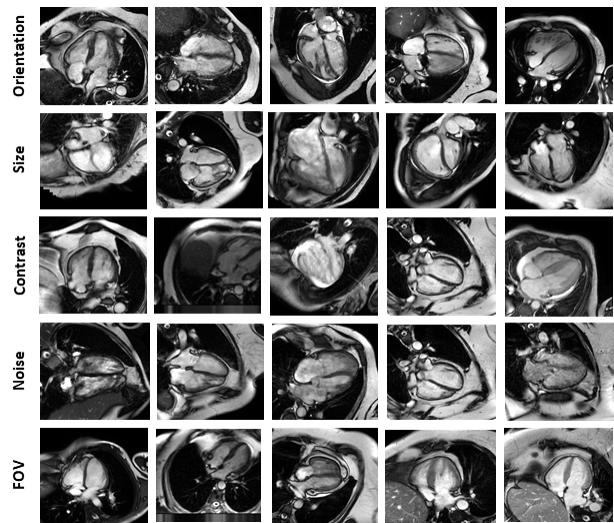}
\caption{Samples from the dataset showing the diversity of the scans with respect to orientation, size, contrast, noise, and field of view (FOV).}
\label{fig:dataset}
\end{center}
\end{figure}

The loss function of the joint networks is as follows:
\begin{align}\label{eq:general_mtl}
\centering
\mathcal{L} &= \LgReg + w_0 \cdot \mathcal{L}_{\mathrm{similarity-D}} + w_1 \cdot \mathcal{L}_{\mathrm{Seg-R}} + w_2 \cdot \mathcal{L}_{\mathrm{Seg-S}},
\end{align}
\begin{align}\label{eq:similarity_D}
\centering
\mathcal{L}_{\mathrm{similarity-D}} &= 1 -  \frac{1}{N}\sum_n \mathrm{LNCC}\left(T_N\circ D_N, D_{temp}\right),
\end{align}
where $\LgReg$ is the GW registration loss defined in Eq. (\ref{eq:groupReg}), $\mathcal{L}_{\mathrm{similarity-D}}$ is the similarity between the warped distance maps and the template distance maps, $\mathcal{L}_{\mathrm{Seg-R}}$ is the segmentation loss between the manual and the estimated contours from propagating the ED contours using the estimated DVF from Group-RegNet, and $\mathcal{L}_{\mathrm{Seg-S}}$ is the segmentation loss between the manual and the predicted contours from SegNet. For more details on the segmentation loss see Section \ref{optimize_SegNet}.  
\section{Dataset, Implementation, and Evaluation}\label{materials_section}

\begin{figure}[t]
\begin{center}
\includegraphics[width=65mm,height=65mm]{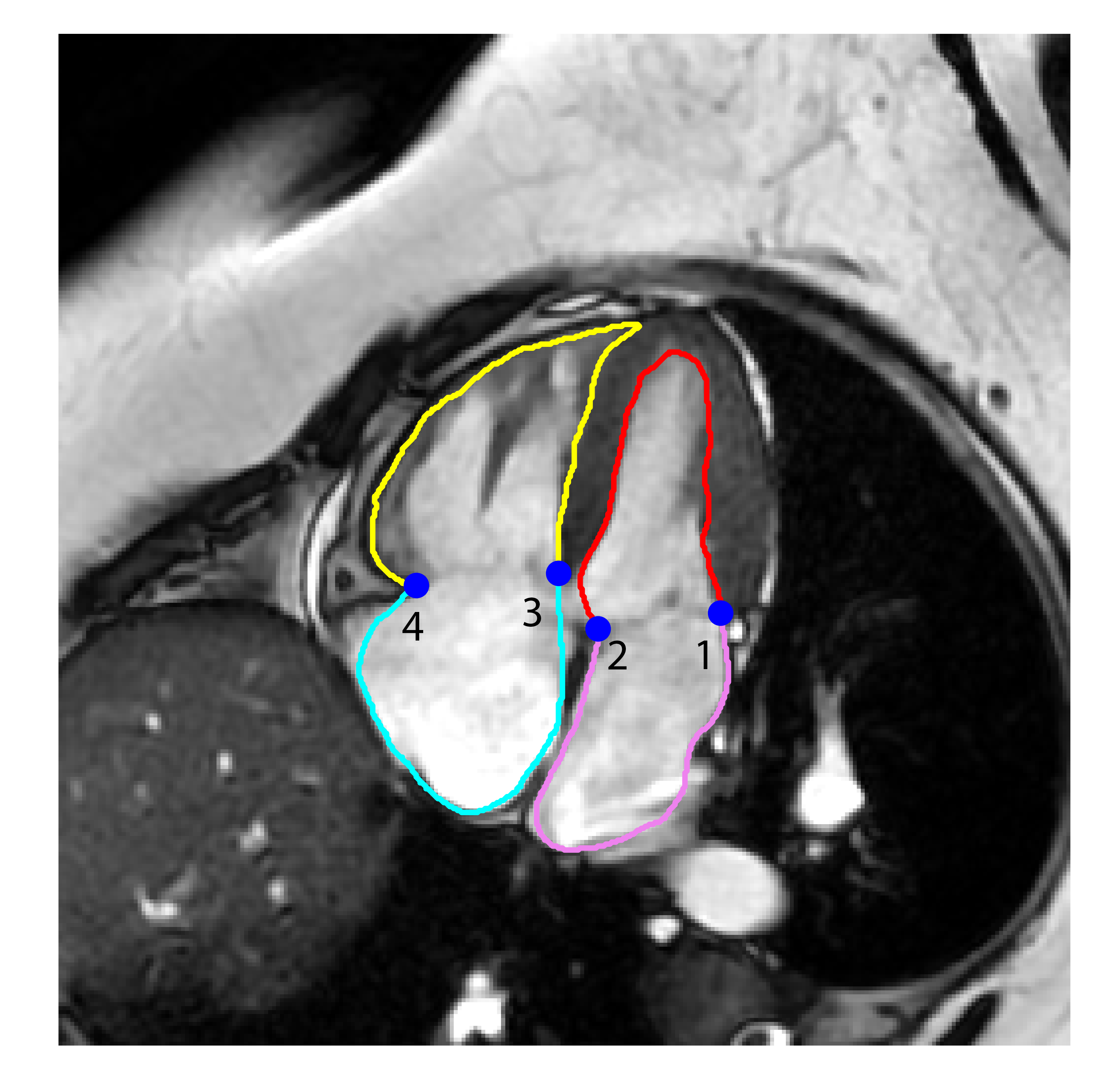}
\caption{Example frame showing contours and landmark points, where the LV, RV, LA, and RA contours are colored in red, yellow, violet, and cyan, respectively.}
\label{fig:landmark}
\end{center}
\end{figure}

\subsection{Dataset} \label{dataset}
This study utilized a cohort of 374 subjects with 2D+t cine-MRI scans, with each subject having a 4-chamber long-axis (LAX) view. The scans were randomly selected from the ASPIRE registry \cite{hurdman2012aspire}, and were acquired using a balanced steady state free precession (bSSFP) sequence. The ASPIRE registry included patients with a wide range of pathologies including left heart disease (15\%), lung disease
(12\%), chronic thromboembolic pulmonary arterial hypertension (PAH) (21\%), PAH (29\%), other PAH (2\%) and non-PAH (21\%). The scans were acquired using Siemens (n=41) and GE scanners (n=333). The dataset was split into 80\% for training (n=300) and 20\% for validation (n=74) scans. \review{The training/validation split was stratified based on the scanner. Specifically, out of the 300 scans used for training, 33 scans were from Siemens, and 267 scans were from GE. For the validation set, consisting of 74 scans, 8 scans were from Siemens, and 66 scans were from GE.} Expert manual contours for the left ventricle (LV), right ventricle (RV), LV myocardium (LVM), left atrium (LA) and right atrium (RA) were defined in all temporal frames by four observers with 2, 3, 11, and 13 years of CMR experience \cite{alandejani2022training}. \review{The number of frames per scan ranged from 20 to 30. Each scan corresponds to a complete cardiac cycle.} The image series and corresponding contours were center-cropped and resampled into image stacks of size $256 \times 256 \times 25$. \review{These images maintain an in-plane resolution of $1 \times 1$ mm, with a total of 25 time frames}. The image stacks were normalized to have intensity values between [0, 1]. A total of four landmark points were defined based on the manual contours, these points represented the hinge points of the mitral and tricuspid valves as shown in Fig. \ref{fig:landmark}. These anatomical locations represent the regions with the largest cardiac motion amplitude and are used for the evaluation of the GW registration. Figure \ref{fig:dataset} illustrates the diversity and variation present in the dataset, as seen through samples depicting a range of field of views, noise levels, contrasts, heart sizes, and orientations.

\subsection{Training and Implementation Details} \label{implementation}
All proposed networks were trained and optimized using PyTorch (version 1.9) with the Adam optimizer and a learning rate of $10^{-4}$. All convolutional layers were initialized using Kaiming initialization. The batch size for the GW registration network was 1 and for the segmentation network 25. The regularization terms $\lambda_0$ and $\lambda_1$ from Eq. (\ref{eq:groupReg}) are empirically set to 0.8 and 0.01. The loss weight w$_0$ from Eq. (\ref{eq:general_mtl}) is set to 5.0, while w$_1$ and w$_2$ are set equally to 1.0 in accordance to the findings from \cite{elmahdy2021joint}. Training was performed on a cluster equipped with NVIDIA RTX6000 cards with 24 GB of memory. % NVIDIA RTX6000, Tesla V100, and GTX1080 Ti GPUs with 24, 16 and 11 GB of memory, respectively. 
To effectively capture the high variance in the dataset, various augmentation techniques were employed, including random mirroring, translation, rotation, scaling, as well as the addition of random Gaussian noise and blurring. For training the proposed CMRINet network, first the SegNet network is trained for 300 epochs, while Group-RegNet is frozen. Afterwards, Group-RegNet is trained for 300 epochs, while SegNet is being frozen. Lastly, both networks are trained simultaneously for 200 epochs. 

\subsection{Evaluation Metrics}
% \subsubsection{Segmentation} 
% The automatically generated contours are evaluated geometrically by comparing them against the manual contours. The Dice similarity coefficient (DSC) measures the overlap between contours:
% \begin{equation}\label{eq:dsc}
% 	\centering
% \mathrm{DSC}= \sum \frac{2 \mid S_m \cap \SGMath \mid}{\mid S_m \mid + \mid \SGMath \mid},
% \end{equation} 
% where $\SGMath$ is the generated contour from either the segmentation or the GW registration network and $S_m$ the manually annotated segmentation. The distance between the contours is measured by the mean contour distance (MCD) and $95\%$ Hausdorff distance (HD) defined as follows:
% \begin{align}
% \centering
% \mathrm{MCD} &= \frac{1}{2} \left( \frac{1}{N} \sum_{i=1}^{n} d \left( a_i, \SGMath \right) +  \frac{1}{M} \sum_{i=1}^{m} d \left( b_i, S_m \right) \right),\label{eq:ch5_msd}
% \\
% \mathrm{HD} &= \max\! \left\lbrace\! \max_i \left\lbrace d \left( a_i, \SGMath \right) \right\rbrace , \max_j \left\lbrace d \left( b_i, S_m \right) \right\rbrace \!\right\rbrace, \label{eq:ch5_hd}
% \end{align}
% where $\{a_1, a_2, \ldots, a_n\}$ and $\{b_1, b_2, \ldots, b_m\}$ are the points of the generated and manual contours, respectively, and $d \left( a_i, \SGMath \right) = \min_j \, \|b_j - a_i\|$.

\subsubsection{Segmentation} 
The automatically generated contours are evaluated geometrically by comparing them to the manual contours. The Dice similarity coefficient (DSC) is used to measure the overlap between the contours, while the mean contour distance (MCD) and Hausdorff distance (HD) measure the distance between the contours. For more details, see \cite{elmahdy2021joint}.
% The automatically generated contours are evaluated geometrically by comparing them to the manual contours using the mean contour distance (MCD) to measure the distance between the contours. For more details, see \cite{elmahdy2021joint}.

\subsubsection{Quality of Deformation Vector Field}
In order to evaluate the quality of the deformations, we calculate the determinant of the Jacobian matrix. 
A determinant of 1 indicates that no volume change has occurred; a determinant $>$ 1 indicates expansion, a determinant between 0 and 1 indicates shrinkage, and a determinant $\leq$ 0 indicates a singularity, i.e. a place where folding has occurred. We can quantify the smoothness and quality of the DVF by indicating the fraction of foldings per image and by calculating the standard deviation of the Jacobian.

\subsubsection{Strain Quantification}
Cardiac longitudinal strain is a non-invasive measure of the deformation or change in length of the myocardial tissue along the longitudinal direction, usually measured in the left ventricle. It is used as an indicator of changes in myocardial contractility and is a sensitive measure of left ventricular function. Strain is typically expressed as a percentage and can range from negative values indicating contraction to positive values indicating relaxation. Factors such as age and gender can affect the strain values. Strain can be measured locally for different segments of the wall or globally across the wall, referred to as global longitudinal strain (GLS). GLS is defined as follows: 
\begin{align}\label{eq:strain}
\centering
GLS = \sum_{i=1}^{n} \epsilon_{i} \;, \hspace{25pt}
\epsilon = \frac{L - L_{0}}{L_{0}},
\end{align}
where $\epsilon_{i}$ is longitudinal strain of the i-th segment, $L$ is the length of the wall at any phase and $L_{0}$ is the length of the wall at the end-diastolic phase.

\section{Experiments and Results} \label{results_section}
\begin{table}[t]
	\centering
	\caption[]{MCD (mm) values for different input frames for SegNet trained with DSC loss. Values in red represent the best results.}
	\resizebox{\linewidth}{!}{
 
        \begin{tabular}{lccccc}
        \hline
         \multirow{2}{*}{Network} & \multicolumn{1}{c}{LV} & \multicolumn{1}{c}{LVM} & \multicolumn{1}{c}{RV} & \multicolumn{1}{c}{LA} & \multicolumn{1}{c}{RA} \\ 
        &\multicolumn{1}{c}{$\mu \pm \sigma$} &\multicolumn{1}{c}{$\mu \pm \sigma$}&\multicolumn{1}{c}{$\mu \pm \sigma$}&\multicolumn{1}{c}{$\mu \pm \sigma$}&\multicolumn{1}{c}{$\mu \pm \sigma$}\\ \hline

        2D & $2.7\pm0.5$ & $2.2\pm0.3$ & $2.8\pm0.5$ & $2.6\pm0.4$ & $\textcolor{red}{3.2\pm0.7}$ \\ \hline
        
        2.5D/1NF & $\textcolor{red}{2.5\pm0.8}$ & $\textcolor{red}{2.0\pm0.5}$ & $\textcolor{red}{2.5\pm0.7}$ & $\textcolor{red}{2.5\pm0.8}$ & $3.9\pm1.0$ \\ \hline

        2.5D/2NF & 2.7$\pm0.6$ & $2.4\pm0.1$ & $2.5\pm0.6$ & $2.6\pm0.2$ & $3.8\pm0.7$\\ \hline
        
        \end{tabular}
        }
\label{table:SegNet_input_MCD}
\end{table}

\begin{table}[t]
	\centering
	\caption[]{MCD (mm) values for different loss functions for SegNet using 2.5D with 1 neighbouring frame. Values in red represent the best results.}
	\resizebox{\linewidth}{!}{
        \begin{tabular}{lccccc}
        \hline
         \multirow{2}{*}{Loss} & \multicolumn{1}{c}{LV} & \multicolumn{1}{c}{LVM} & \multicolumn{1}{c}{RV} & \multicolumn{1}{c}{LA} & \multicolumn{1}{c}{RA} \\ 
        &\multicolumn{1}{c}{$\mu \pm \sigma$} &\multicolumn{1}{c}{$\mu \pm \sigma$}&\multicolumn{1}{c}{$\mu \pm \sigma$}&\multicolumn{1}{c}{$\mu \pm \sigma$}&\multicolumn{1}{c}{$\mu \pm \sigma$}\\ \hline
        
        DSC & $2.5\pm0.8$ & $2.0\pm0.5$ & $2.5\pm0.7$ & $2.5\pm0.8$ & $3.9\pm1.0$ \\ \hline
        
        CE & $2.0\pm0.6$ & $1.6\pm0.4$ & $2.1\pm0.6$ & $3.0\pm1.1$ & $3.6\pm1.2$ \\ \hline
        
        DSC+CE & $\textcolor{red}{1.8\pm0.5}$ & $\textcolor{red}{1.5\pm0.4}$ & $\textcolor{red}{2.0\pm0.5}$ & $\textcolor{red}{2.1\pm0.8}$ & $\textcolor{red}{2.8\pm1.0}$\\ \hline
        
        \end{tabular}
        }
\label{table:SegNet_loss_MCD}
\end{table}
In this paper, we present an end-to-end fully automatic joint segmentation and GW registration network dubbed CMRINet. The network comprised of two networks for segmentation and GW registration namely, SegNet and Group-RegNet (See Sections \ref{segNetwork}, \ref{groupRegNetwork}).
Following, we optimize SegNet and Group-RegNet on the validation set before jointly optimizing them for CMRINet. 
% In this paper we present two base networks dubbed SegNet and group-RegNet for segmentation and groupwise registration, respectively. We also investigated joining these two tasks via different network architectures, namely \textcolor{red}{xx, xx, xx, xx} (see Sections \ref{segNetwork}, \ref{groupRegNetwork}, \ref{JointNetwork}). We compared our proposed networks against a well-established open source image registration toolbox, named \texttt{elastix} \cite{elastix} \cite{metz2011nonrigid}. \texttt{Elastix} uses conventional iterative registration. We used NCC similarity loss using the settings proposed by \cite{shahzad2017fully}
% In this paper, we introduce two base networks: "Seg" for segmentation and "groupReg" for groupwise registration. We also explore different network architectures for jointly performing these tasks (refer to Sections \ref{segNetwork}, \ref{groupRegNetwork}, and \ref{JointNetwork}). We compare the performance of our proposed networks with an established open-source toolbox, "elastix" \cite{elastix}, which utilizes conventional iterative registration.
% Then, we present the final results compared to literature on the independent test set \textcolor{red}{??}.

\subsection{Optimizing the Segmentation Network} \label{optimize_SegNet}
The cine-MRI scans in the dataset have an anisotropic resolution, however they are continuous in time and in order to ensure continuity in the network predictions, we hypothesize that stacking neighboring frames could be beneficial. To validate this assumption, we conducted an experiment to determine whether including neighboring frames in the segmentation network would be beneficial. We compared a 2D training scheme (using only the center frame) with two alternative 2.5D schemes: (i) using the direct neighboring frames together with the center frame as input (network 2.5D/1NF) and (ii) using two neighboring frames with the center frame (network 2.5D/2NF). In both cases, only the center frame was segmented and used in the loss function. The DSC loss function was utilized in this experiment. The results of this experiment in terms of MCD are presented in Table \ref{table:SegNet_input_MCD}, which demonstrates that the 2.5D scheme with one neighboring frame consistently outperforms the 2D and 2.5D/2NF training schemes. Therefore, we selected this scheme for the following experiments.

In the next experiment, we examined the impact of different loss functions on the network performance. We tested three loss functions, namely DSC, cross entropy (CE), and the equal sum of both (DSC+CE). The performance of the network with different loss functions is presented in Table \ref{table:SegNet_loss_MCD}. The combined DSC and CE loss yielded the best results, and we therefore adopted this loss function for the final model.

\begin{table}[t]
	\centering
	\caption[]{MCD (mm) values for different template estimation methods for Group-RegNet. Values in red represent the best results.}
	\resizebox{\linewidth}{!}{
 
        \begin{tabular}{lccccc}
        \hline
         \multirow{2}{*}{Template} & \multicolumn{1}{c}{LV} & \multicolumn{1}{c}{LVM} & \multicolumn{1}{c}{RV} & \multicolumn{1}{c}{LA} & \multicolumn{1}{c}{RA} \\ 
        &\multicolumn{1}{c}{$\mu \pm \sigma$} &\multicolumn{1}{c}{$\mu \pm \sigma$}&\multicolumn{1}{c}{$\mu \pm \sigma$}&\multicolumn{1}{c}{$\mu \pm \sigma$}&\multicolumn{1}{c}{$\mu \pm \sigma$}\\ \hline
        
        Average & $\textcolor{red}{1.7\pm0.8}$ & $\textcolor{red}{1.3\pm0.5}$ & $\textcolor{red}{1.6\pm0.7}$ & $\textcolor{red}{1.3\pm0.5}$ & $\textcolor{red}{1.8\pm0.7}$ \\ \hline
        
        PCA & $1.8\pm0.9$ & $1.4\pm0.5$ & $1.9\pm0.8$ & $1.4\pm0.5$ & $2.0\pm0.8$ \\ \hline
        
        \end{tabular}
        }
\label{table:template}
\end{table}

\subsection{Optimizing the Groupwise Registration Network}
One of the main components of GW registration algorithms is the template estimation approach. Several template selection approaches have been demonstrated to be effective \cite{jia2011intermediate}. These methods include i) the selection of one of the images in the series to be the template, ii) using the average of the warped images as a template, or iii) using Principle Component Analysis (PCA) to estimate the principle image in the series. In order to evaluate which template estimation approach is best suited for the underlying problem we experimented with the average template estimation as shown in Eq. (\ref{eq:template}) and PCA template estimation. The PCA template estimation was inspired by the work from \cite{che2019deep}, where the template image is constructed using the eigenvector $V = \{ v_n|n=1, \ldots, N\}$, which corresponds to the highest eigenvalue and serves as the weights for the warped images as shown in the following equation:
\begin{align}\label{eq:template}
\centering
I_{temp} = \sum_{n=1}^{N} v_n (T_n\circ I_n),
\end{align} 

Table \ref{table:template} shows the performance of different template estimation methods. The average template results demonstrated better performance compared to the PCA template. Therefore, we selected the average template method for Group-RegNet network.

\begin{table}[t]
	\centering
	\caption[]{MCD (mm) values for different CMRINet configurations. Values in red represent the best results.}
	\resizebox{\linewidth}{!}{
 
        \begin{tabular}{lccccc}
        \hline
         \multirow{2}{*}{Input} & \multicolumn{1}{c}{LV} & \multicolumn{1}{c}{LVM} & \multicolumn{1}{c}{RV} & \multicolumn{1}{c}{LA} & \multicolumn{1}{c}{RA} \\ 
        &\multicolumn{1}{c}{$\mu \pm \sigma$} &\multicolumn{1}{c}{$\mu \pm \sigma$}&\multicolumn{1}{c}{$\mu \pm \sigma$}&\multicolumn{1}{c}{$\mu \pm \sigma$}&\multicolumn{1}{c}{$\mu \pm \sigma$}\\ \hline

         % All2All& $1.95\pm0.5$ & $1.7\pm0.3$ & $1.8\pm0.4$ & $1.9\pm0.3$ & $2.8\pm0.4$\\ \hline

         CMRINet$^a$ & $1.8\pm0.9$ & $1.2\pm0.4$ & $1.2\pm0.3$ & \textcolor{red}{$1.0\pm0.4$} & $1.2\pm0.4$ \\ \hline

         CMRINet$^b$ & $1.7\pm0.8$ & $1.1\pm0.4$ & $1.2\pm0.4$ & $1.1\pm0.3$ & $1.2\pm0.4$ \\ \hline

         Proposed & \textcolor{red}{$1.7\pm0.3$} & \textcolor{red}{$1.0\pm0.1$} & \textcolor{red}{$1.1\pm0.3$} & $1.1\pm0.1$ & \textcolor{red}{$1.1\pm0.2$} \\ \hline
        
        \end{tabular}
        }
\label{table:joint_MCD}
\end{table}

\begin{table*}[t]
	\centering
	\caption[]{MCD (mm) values for \review{the registration output of different methods}. The Mean column  represents the average across all frames, while ES represents the average for the ES frames only. x$\pm$x represents mean$\pm$std. Values in red represent the best results. \review{Daggers denote one-way ANOVA statistical significance for the proposed network against other networks.}}
	\resizebox{\linewidth}{!}{
 
        \begin{tabular}{lcc|cc|cc|cc|cc}
        \hline
         \multirow{2}{*}{Network} & \multicolumn{2}{c}{LV} & \multicolumn{2}{c}{LVM} & \multicolumn{2}{c}{RV} & \multicolumn{2}{c}{LA} & \multicolumn{2}{c}{RA} \\ 
        
        % &\multicolumn{2}{c}{$\mu \pm \sigma$} &\multicolumn{2}{c}{$\mu \pm \sigma$}&\multicolumn{2}{c}{$\mu \pm \sigma$}&\multicolumn{2}{c}{$\mu \pm \sigma$}&\multicolumn{2}{c}{$\mu \pm \sigma$}\\ 

        & Mean & ES &  Mean & ES &  Mean & ES &  Mean & ES &  Mean & ES \\ \hline 
        
        % w/o Reg. & $3.5\pm2.1$ & $6.3 \pm 1.7$ & $2.2\pm1.1$ & $3.7 \pm 0.9$ & $2.9\pm1.5$ & $4.9 \pm 1.4$ & $2.1\pm1.0$ & $3.4 \pm 1.2$ & $2.9\pm1.4$ & $4.7 \pm 1.9$ \\ \hline
        
        \texttt{elastix}& \textcolor{red}{$1.7\pm0.7$} & \textcolor{red}{$2.5 \pm 1.0$} &$1.3\pm0.5$ & $1.9 \pm 0.6$ & $1.8\pm0.8$ & $2.6 \pm 1.2$ & $1.3\pm0.5$ & $1.6 \pm 0.8$ & $1.8\pm0.8$ & $2.5 \pm 1.3$ \\ \hline 

        Group-RegNet & $1.7\pm0.8$ & $2.6 \pm 0.9$ & $1.3\pm0.5$ & $1.9 \pm 0.6$ & $1.6\pm0.7$ & $2.3 \pm 1.0$ & $1.3\pm0.5$ & $1.6 \pm 0.7$ & $1.8\pm0.7$ & $2.3 \pm 1.1$  \\ \hline

        Dense & $1.8\pm0.8$ & $2.8 \pm 1.3$  & $1.3\pm0.5$ & $1.8 \pm 0.9$  & $1.4\pm0.5$ & $1.9 \pm 0.9$ & $1.1\pm0.4$ & $1.4 \pm 0.5$  & $1.4\pm0.4$ & $1.8 \pm 0.8$  \\ \hline  

        SEDD & $1.8\pm0.7$ & $2.5 \pm 1.2$   & $1.4\pm0.5$ & $1.7 \pm 0.8$   & $1.5\pm0.7$ & $1.9 \pm 0.8$ & $1.3\pm0.5$ & $1.7 \pm 0.5$  & $1.6\pm0.7$ & $1.7 \pm 0.7$  \\ \hline

        Proposed & $1.7\pm0.8$ & $2.6 \pm 1.3$ & \textcolor{red}{$1.1\pm0.4^{\dagger}$} & \textcolor{red}{$1.5 \pm 0.8^{\dagger}$} & \textcolor{red}{$1.2\pm0.4^{\dagger}$} & \textcolor{red}{$1.5 \pm 0.8^{\dagger}$} & \textcolor{red}{$1.1\pm0.3$} & \textcolor{red}{$1.3 \pm 0.6$} & \textcolor{red}{$1.2\pm0.4^{\dagger}$} & \textcolor{red}{$1.4 \pm 0.5^{\dagger}$} \\ \hline
        
        \end{tabular}
        }
\label{table:comaprison_MCD}
\end{table*}

\begin{table*}[t]
	\centering
	\caption[]{MCD (mm) values for landmark points for \review{the registration output of different methods}. The Mean column  represents the average across all frames, while ES represents the average for the ES frames only. x$\pm$x represents mean$\pm$std. Values in red represent the best results. \review{Daggers denote one-way ANOVA statistical significance for the proposed network against other networks.}}
	% \resizebox{\linewidth}{!}{
 
        \begin{tabular}{lcc|cc|cc|cc}
        \hline
         \multirow{2}{*}{Network} & \multicolumn{2}{c}{Landmark 1} & \multicolumn{2}{c}{Landmark 2} & \multicolumn{2}{c}{Landmark 3} & \multicolumn{2}{c}{Landmark 4}  \\ 
        
        % &\multicolumn{2}{c}{$\mu \pm \sigma$} &\multicolumn{2}{c}{$\mu \pm \sigma$}&\multicolumn{2}{c}{$\mu \pm \sigma$}&\multicolumn{2}{c}{$\mu \pm \sigma$}&\multicolumn{2}{c}{$\mu \pm \sigma$}\\ 

        & Mean & ES &  Mean & ES &  Mean & ES &  Mean & ES  \\ \hline 
        
        % w/o Reg. & $xx\pm xx$  & $xx\pm xx$  & $xx\pm xx$  & $xx\pm xx$  & $xx\pm xx$  & $xx\pm xx$  & $xx\pm xx$  & $xx\pm xx$     \\ \hline
        
        \texttt{elastix}& $2.3\pm 1.1$  & $2.8\pm 1.7$  & $2.9\pm 2.0$  & $4.0\pm 3.3$  & $3.2\pm 1.8$  & $4.2\pm 2.8$  & $5.1\pm 3.0$  & $7.3\pm 6.1$   \\ \hline 

        Group-RegNet & $2.3\pm 1.2$  & $2.7\pm 1.9$  & $2.9\pm 2.1$  & $3.9\pm 3.3$  & $3.2\pm 1.7$  & $3.9\pm 2.9$  & $5.1\pm 3.3$  & $7.1\pm 6.1$    \\ \hline

        Dense& $2.6\pm 1.2$  & $3.4\pm 2.6$  & $3.1\pm 2.1$  & $4.5\pm 3.4$  & $3.6\pm 1.9$  & $4.8\pm 3.6$  & $3.2\pm 2.1$  & $8.3\pm 6.2$  \\ \hline  

        SEDD & $2.3\pm 1.2$  & $2.8\pm 1.9$  & $2.9\pm 2.0$  & $4.1\pm 3.3$  & $3.3\pm 1.8$  & $4.1\pm 3.0$  & $5.1\pm 3.3$  & $7.5\pm 6.3$   \\ \hline

        Proposed & \textcolor{red}{$2.2\pm 1.2$}  & \textcolor{red}{$2.7\pm 2.1$}  & \textcolor{red}{$2.6\pm 1.9^{\dagger}$}  & \textcolor{red}{$3.4\pm 3.1^{\dagger}$}  & \textcolor{red}{$3.0\pm 1.5^{\dagger}$}  & \textcolor{red}{$3.6\pm 2.2^{\dagger}$}  & \textcolor{red}{$4.1\pm 2.8$}  & \textcolor{red}{$5.4\pm 4.7$}  \\ \hline
        
        \end{tabular}
        % }
\label{table:landmark_MCD}
\end{table*}

\begin{table}[t]
\centering
\caption{Analysis of the determinant of the Jacobian. Values in red represent the best results. \review{Daggers denote one-way ANOVA statistical significance for the proposed network against other networks.}}
\label{table:dvf}
\begin{tabular}{lcc}
\hline

\multirow{2}{*}{Network} & Std. Jacobian & Folding fraction \\ 
& \multicolumn{1}{c}{$\mu \pm \sigma$} & \multicolumn{1}{c}{$\mu \pm \sigma$} \\ \hline

\texttt{elastix} & $0.17\pm 0.12$ & $0.005\pm 0.002$ \\ \hline
Group-RegNet & $0.18\pm 0.10$ & $0.005\pm 0.003$ \\ \hline
Dense  & $0.17 \pm 0.06$ & $0.005\pm 0.001$ \\\hline
SEDD  & $0.17\pm 0.05$ & $0.003\pm 0.001$ \\\hline
Proposed  & \textcolor{red}{$0.11\pm 0.02^{\dagger}$} & \textcolor{red}{$0.001\pm 0.002^{\dagger}$} \\\hline

\end{tabular}
\end{table}

\begin{figure*}[t]
	\centering
	\resizebox{\textwidth}{!}{
		\begin{tabular}{cccc} 
			 \small{Manual} & \small{Prediction} & \small{DVF} & \small{Jacobian det.}\\

               \includegraphics[width=40mm,height=40mm]{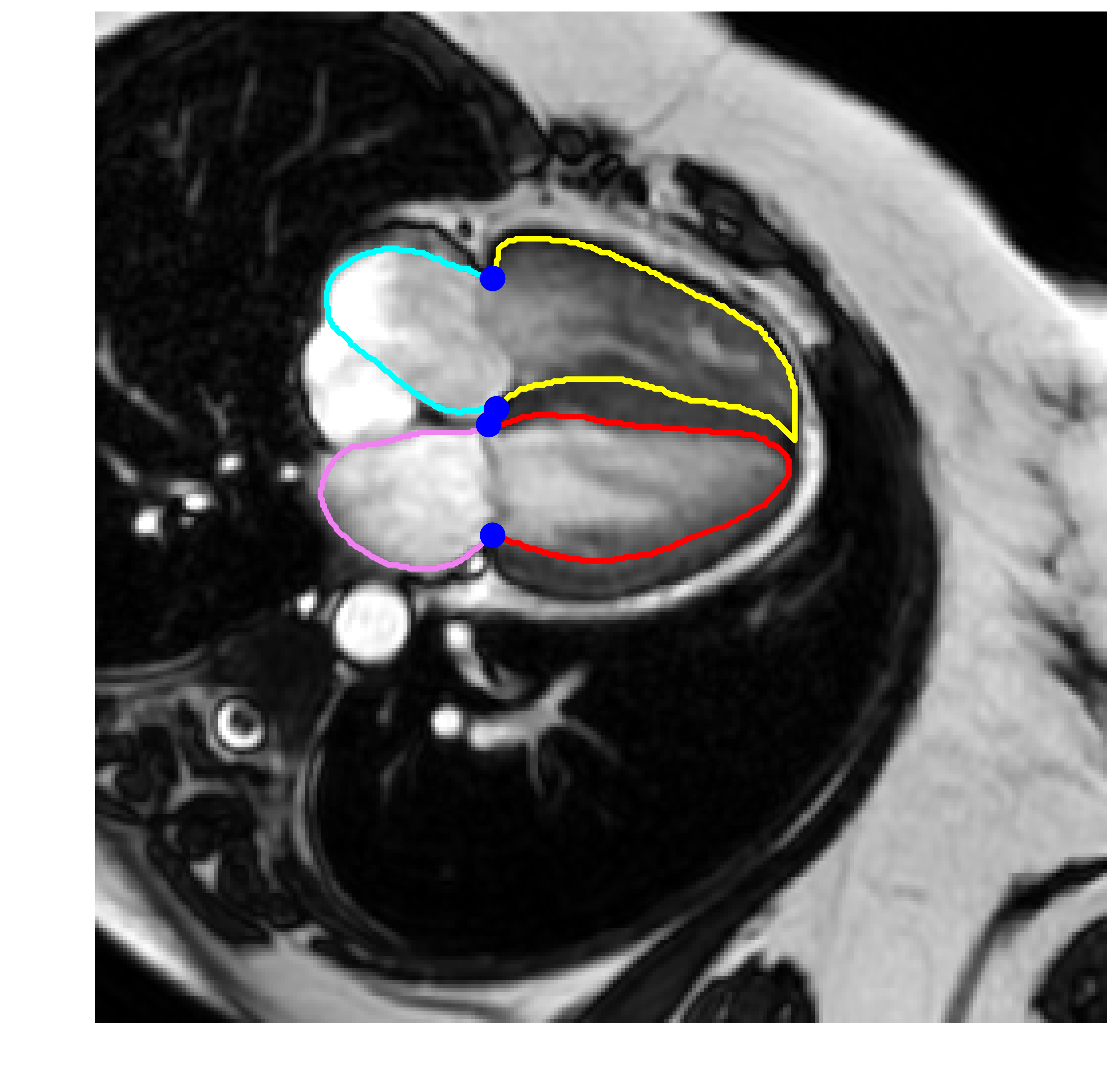} &
               \includegraphics[width=40mm,height=40mm]{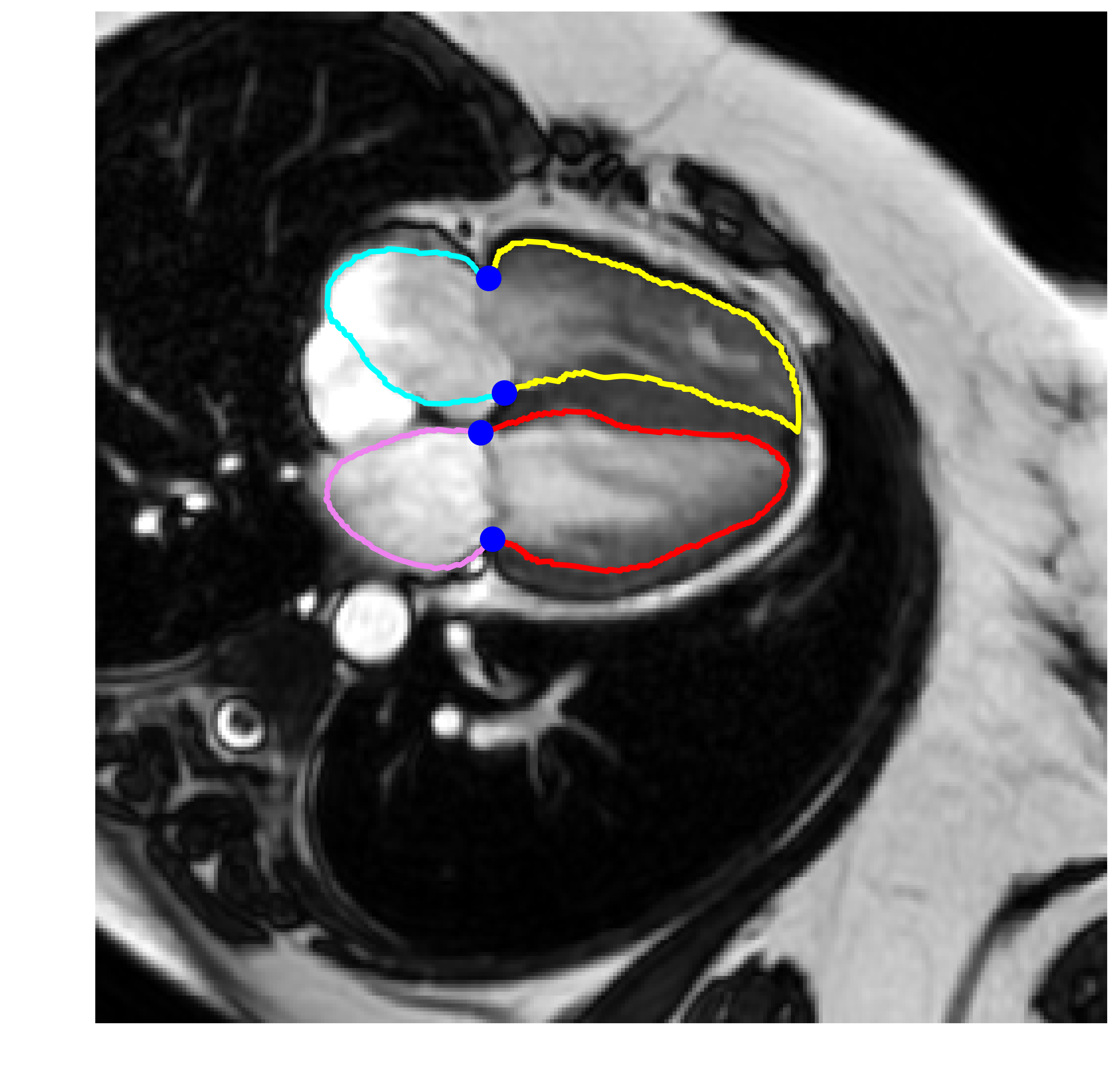}&
               \includegraphics[width=40mm,height=40mm]{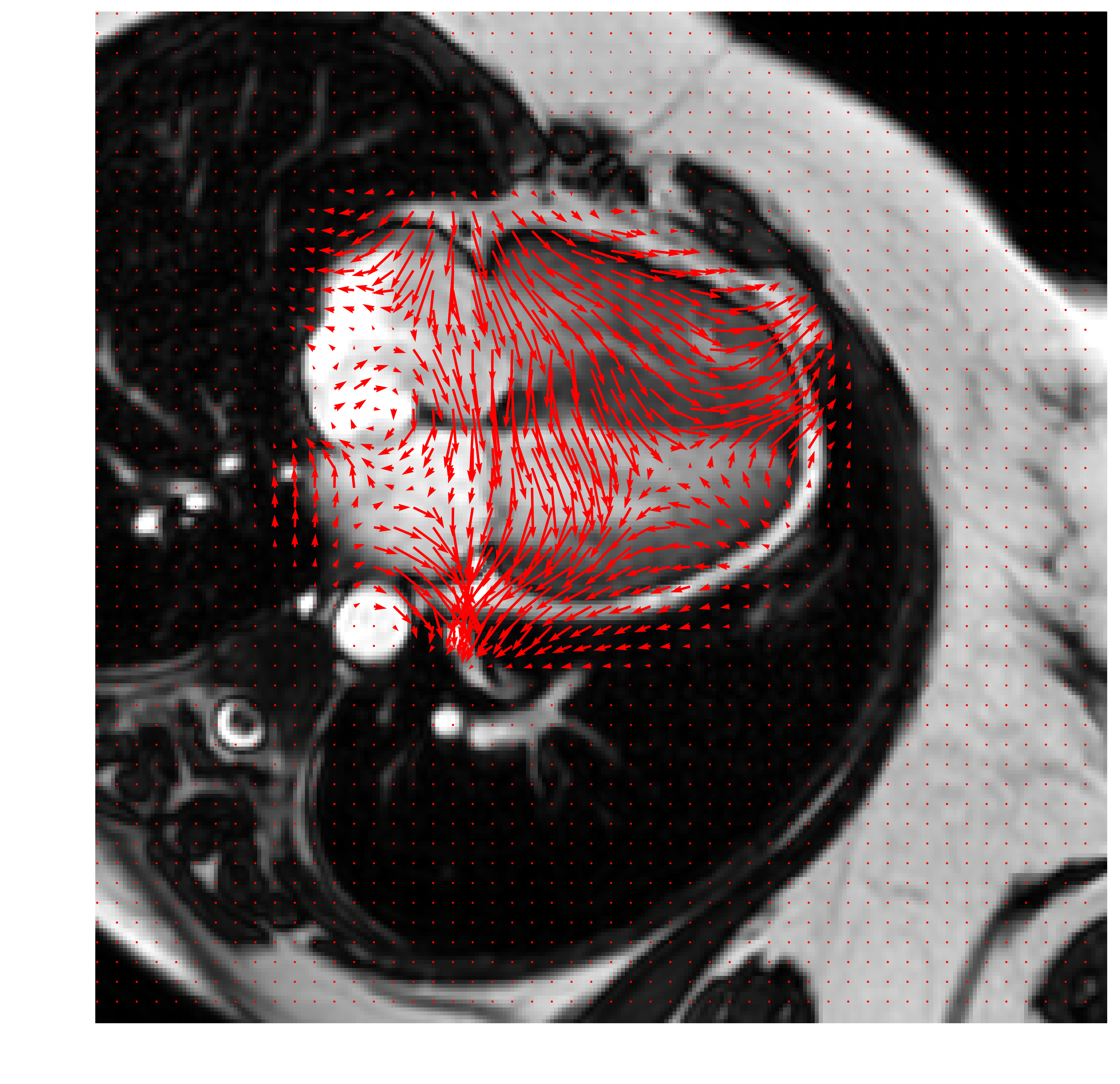}&
               \includegraphics[width=40mm,height=40mm]{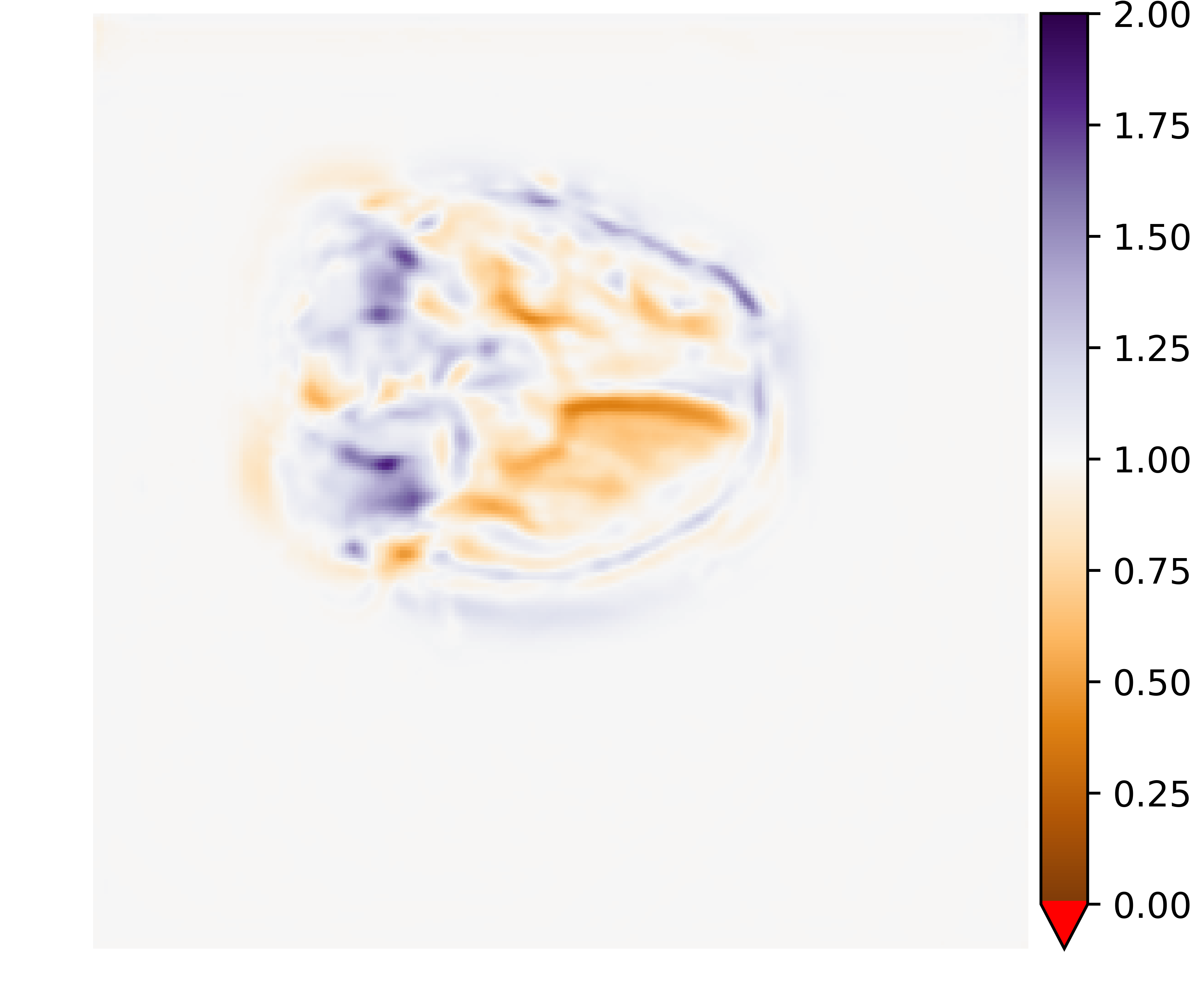} \\

               \includegraphics[width=40mm,height=40mm]{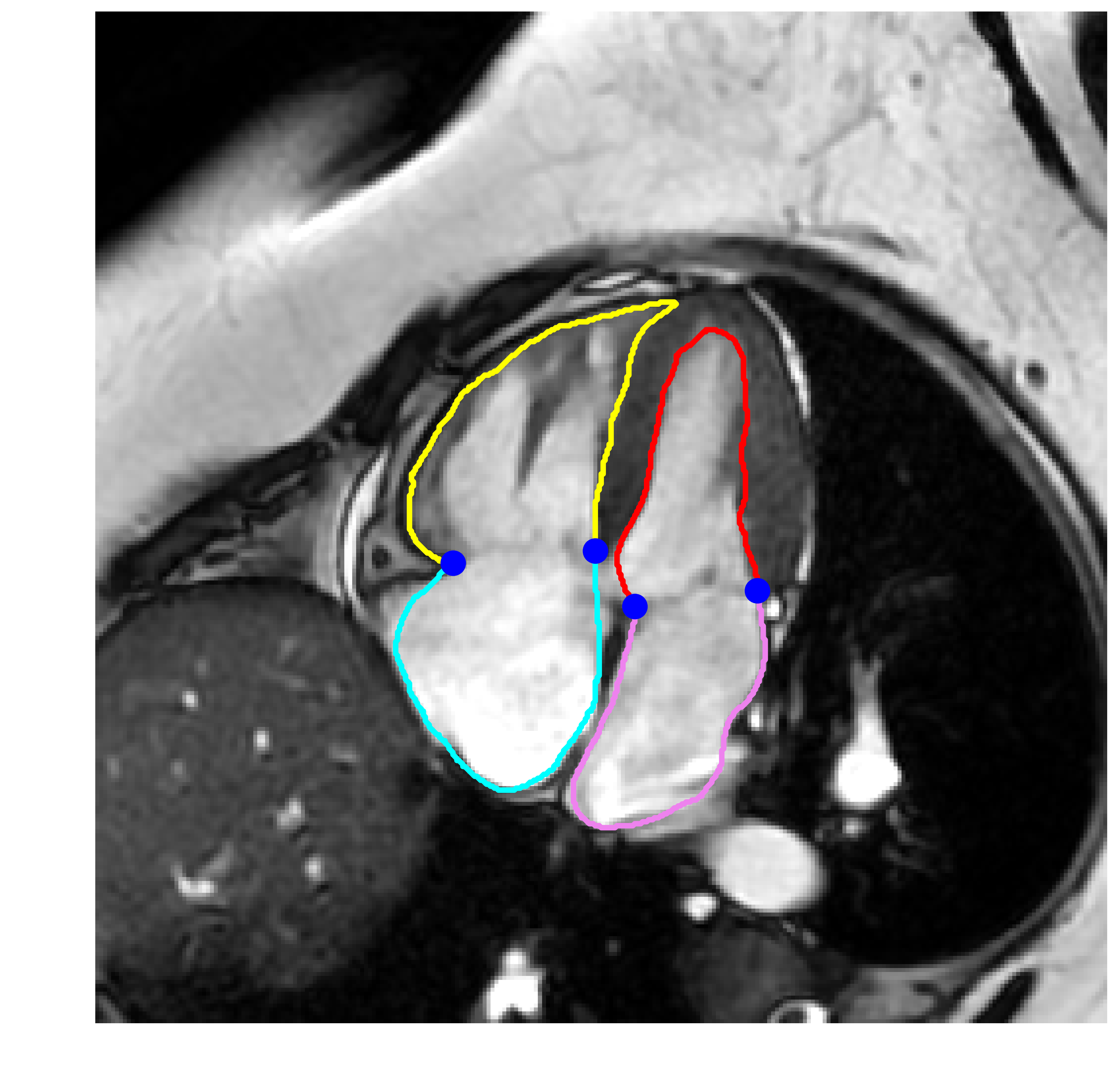} &
               \includegraphics[width=40mm,height=40mm]{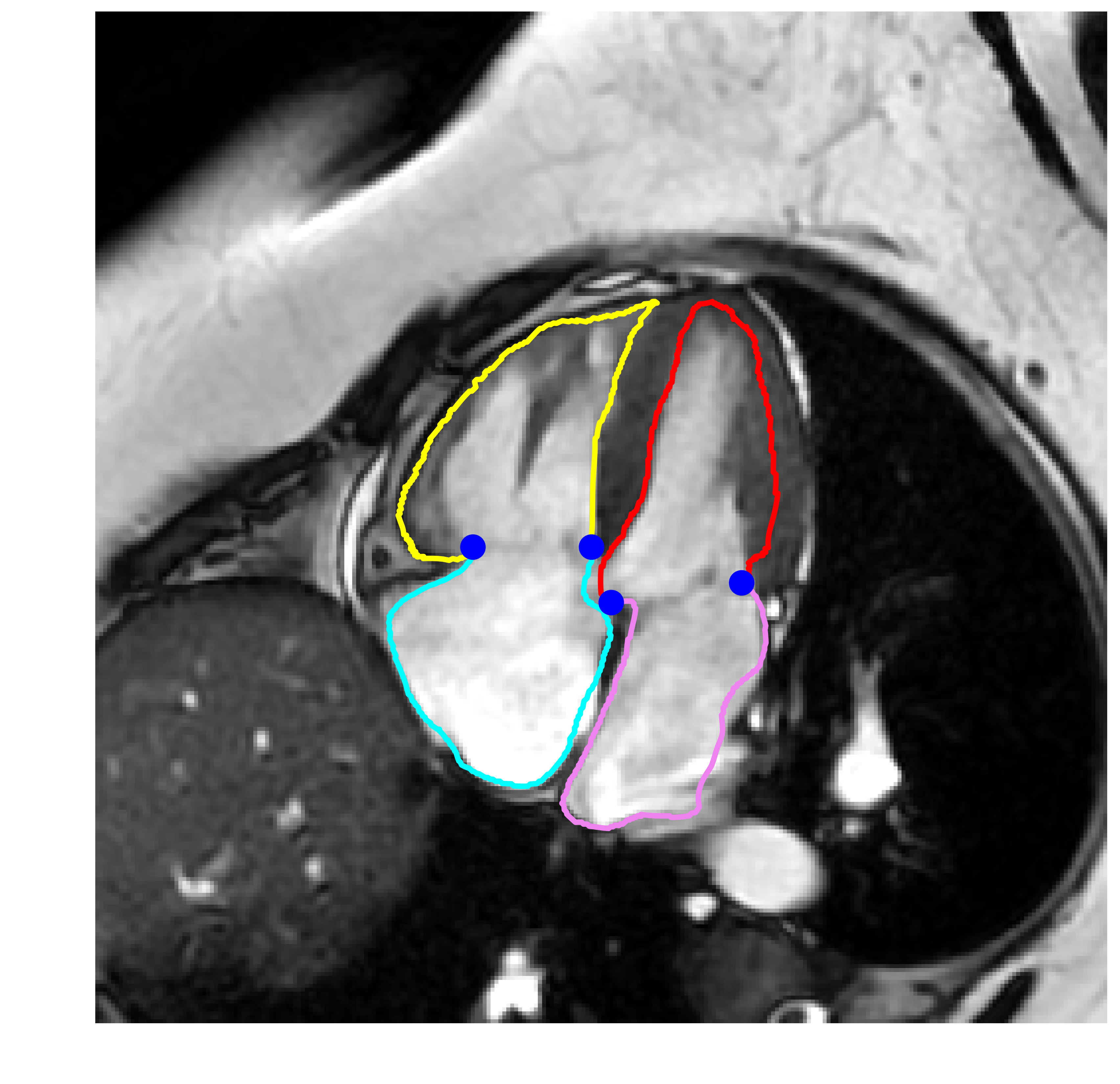}&
               \includegraphics[width=40mm,height=40mm]{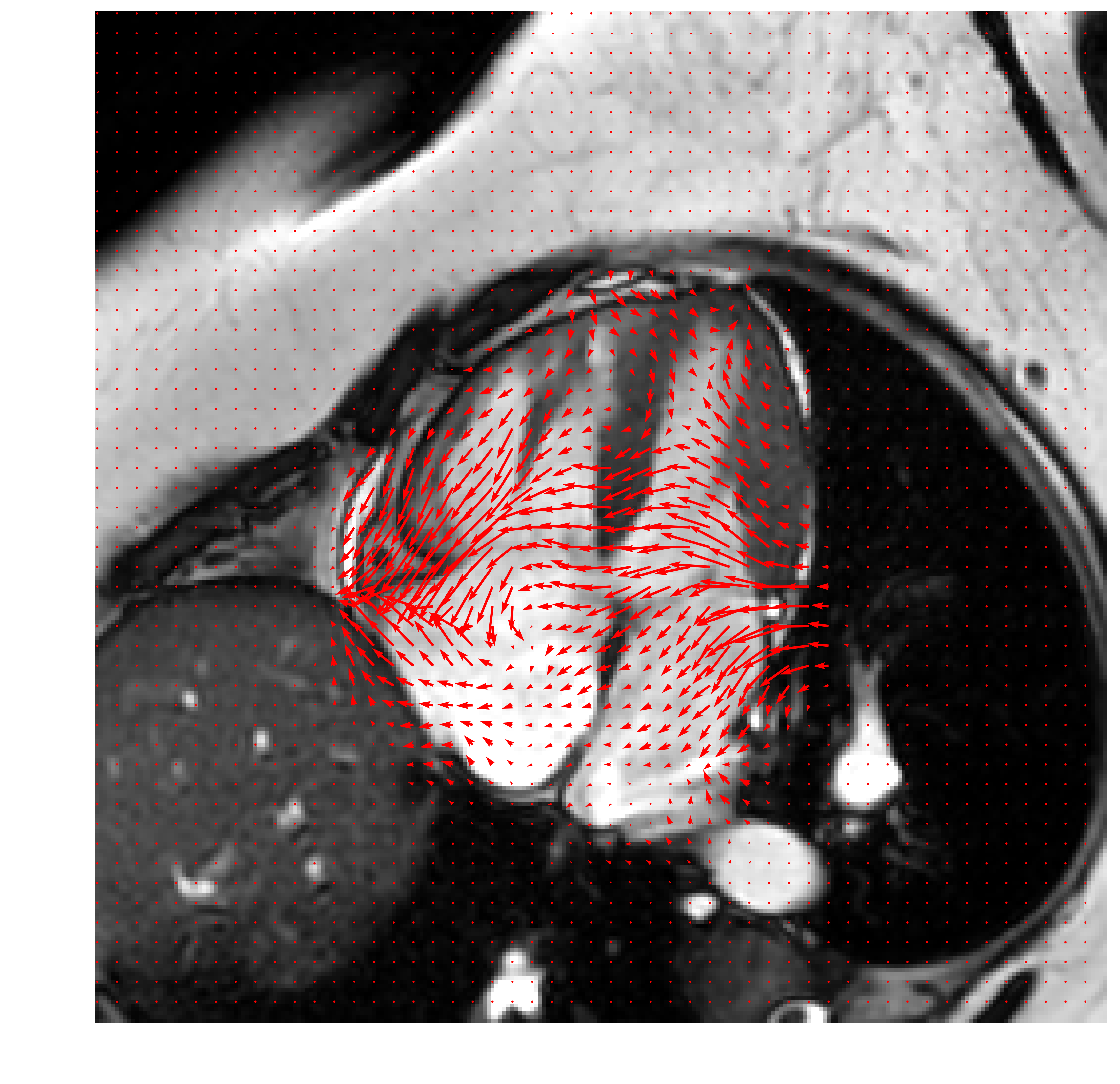}&
               \includegraphics[width=40mm,height=40mm]{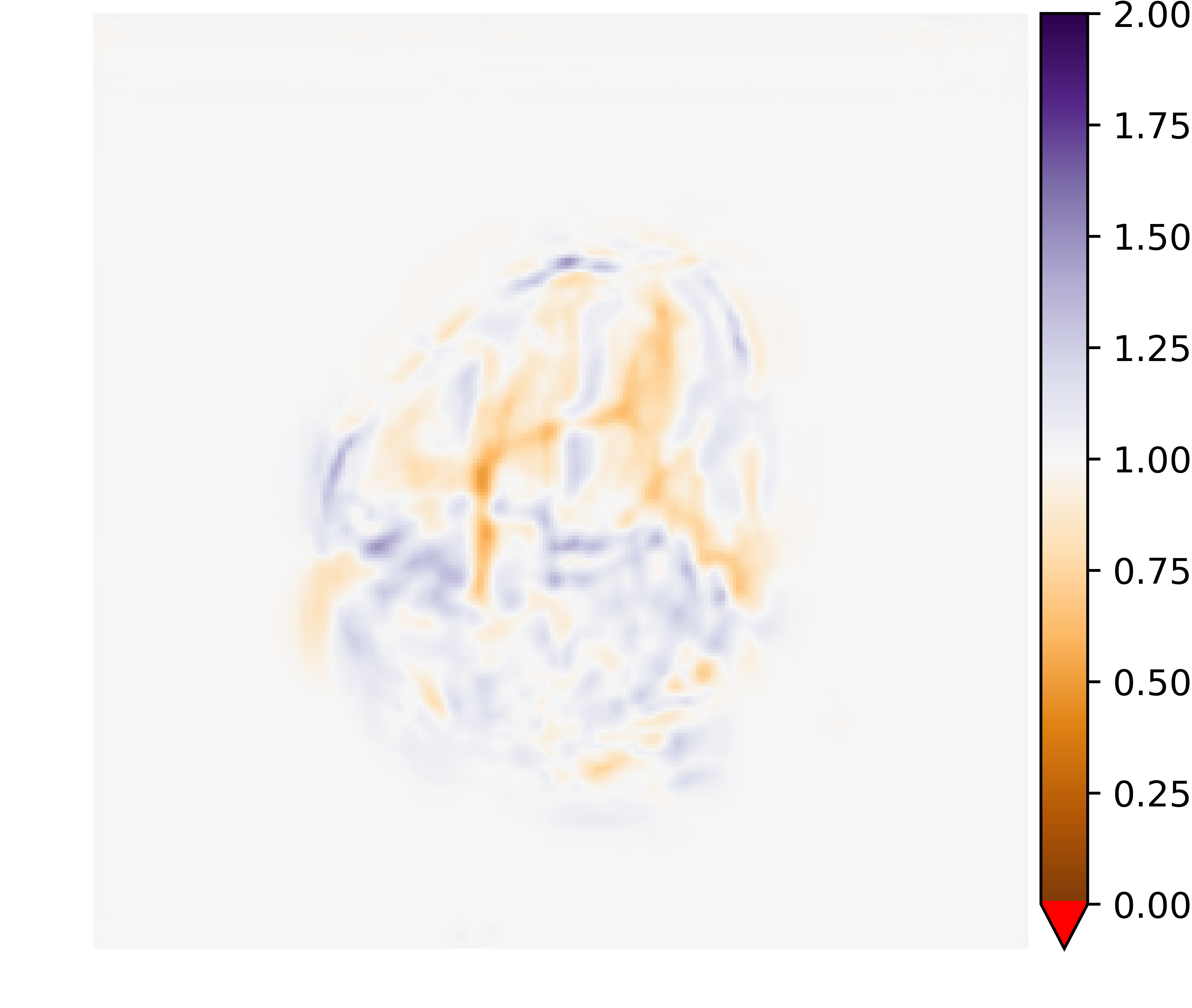} \\

               \includegraphics[width=40mm,height=40mm]{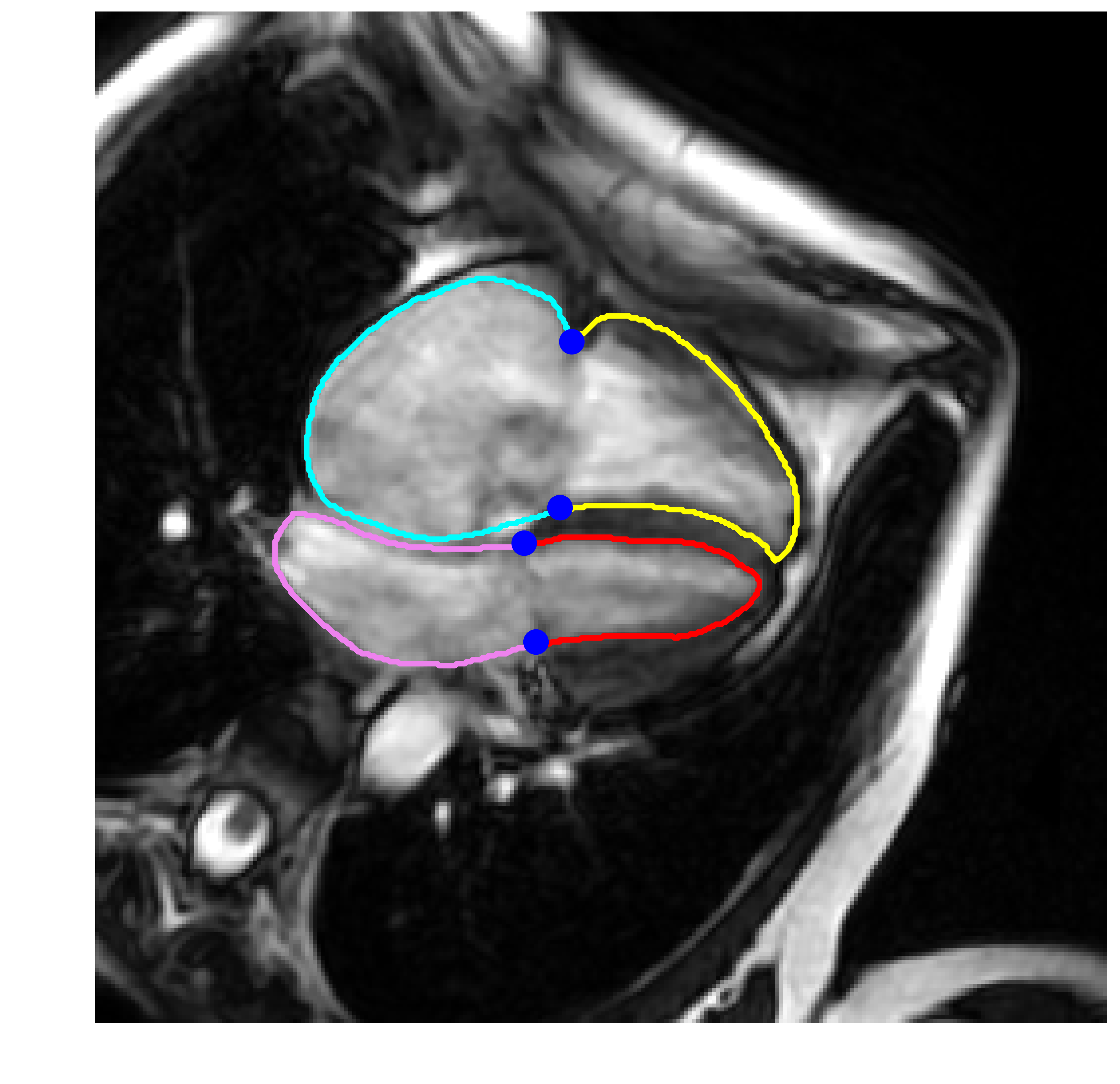} &
               \includegraphics[width=40mm,height=40mm]{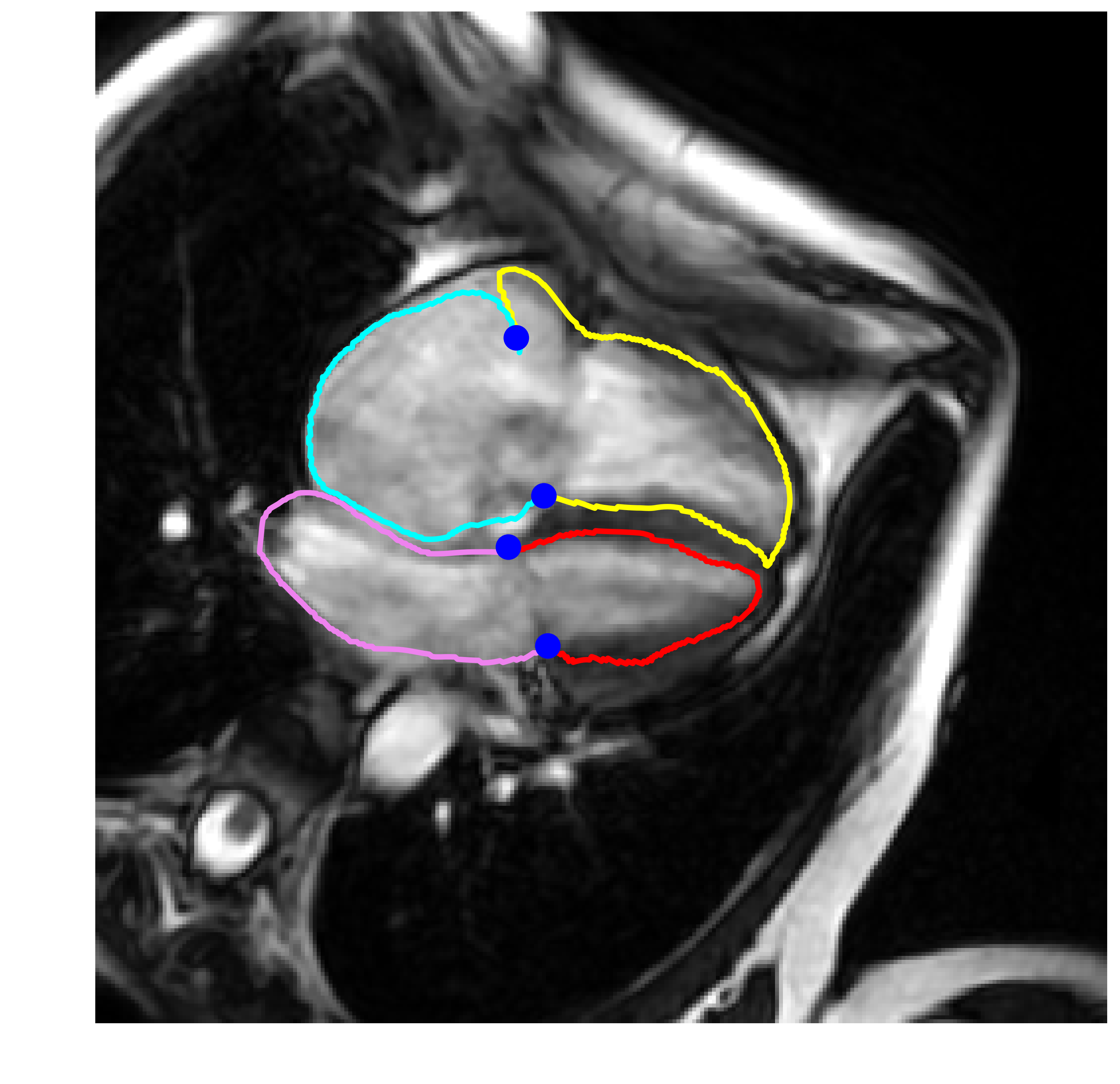}&
               \includegraphics[width=40mm,height=40mm]{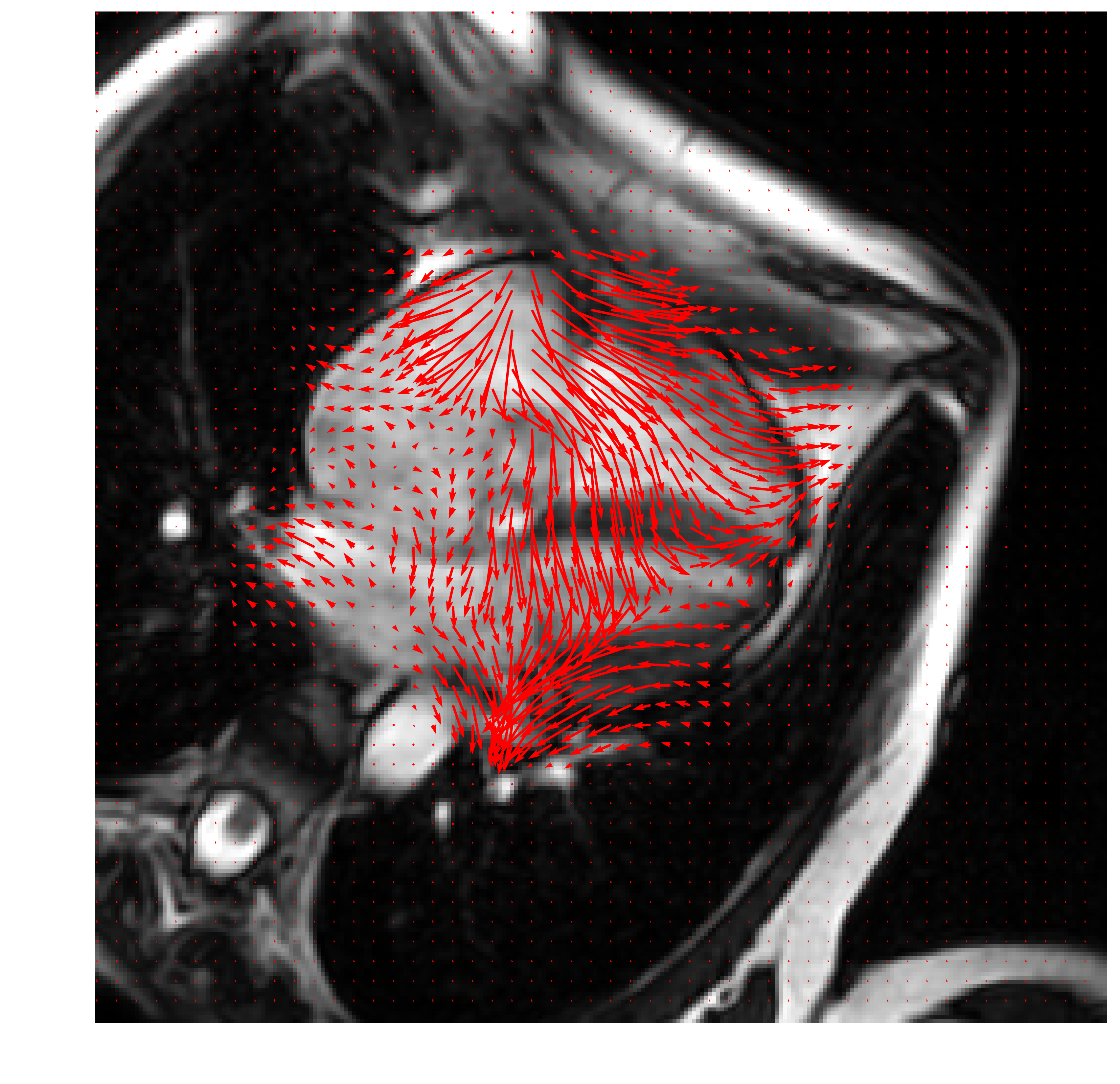}&
               \includegraphics[width=40mm,height=40mm]{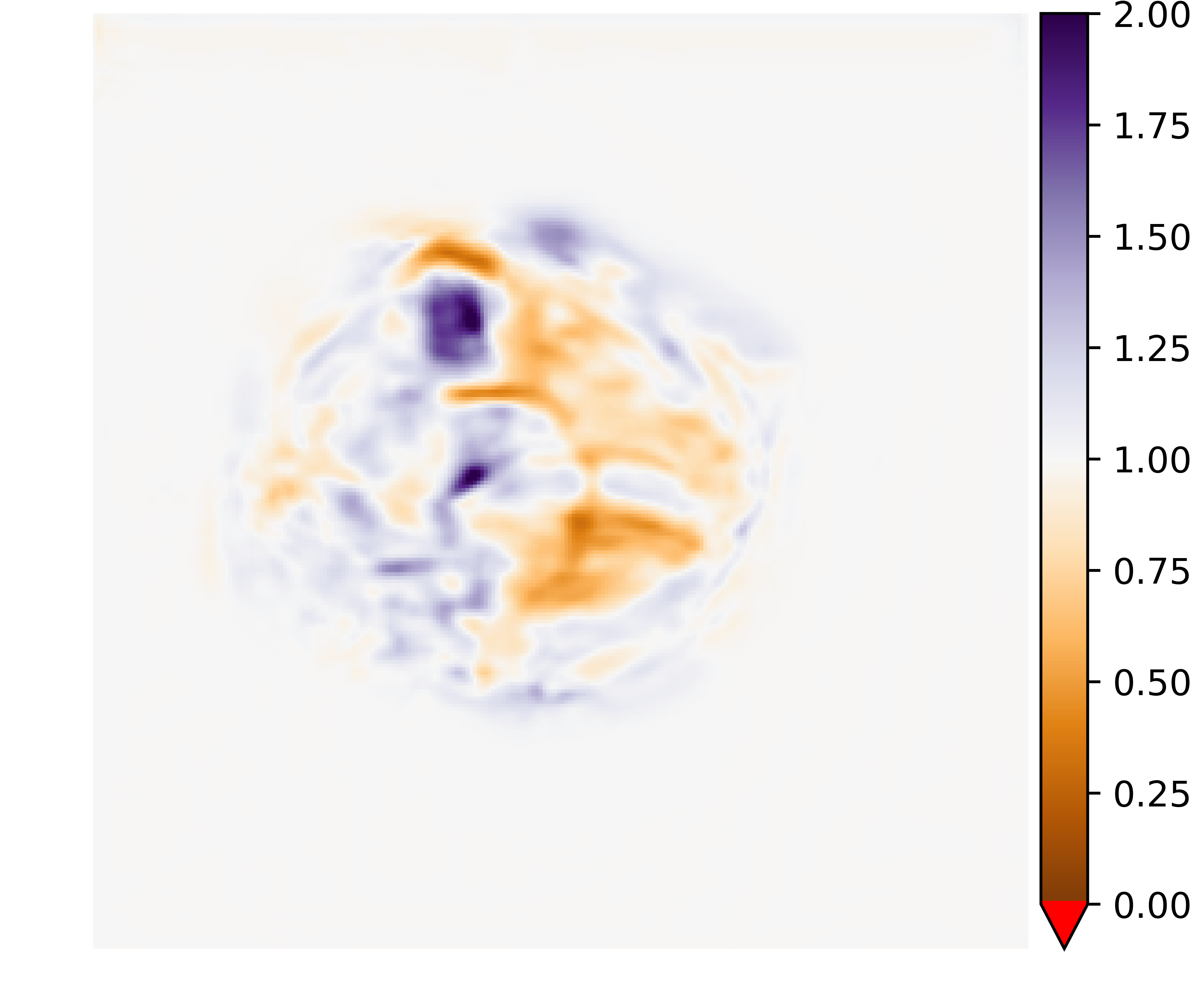} \\
			
		\end{tabular}
	}
\caption{Examples of the ES frame of the proposed method. From top to bottom, the selected cases are the first, second, and third quartile in terms of the MCD of the LV. From left to right, manually defined contours, propagated contours from the automatically segmented ED frame, the DVF quiver, and the determinant of the Jacobian. The LV, RV, LA, and RA contours are colored in red, yellow, violet, and cyan, respectively. Additionally, the landmark points are marked in blue.}
\label{fig:proposed_examples}
\end{figure*}

\begin{figure*}[t]
	\centering
	\resizebox{\textwidth}{!}{
		\begin{tabular}{cccc} 
			 \small{Manual} & \small{Elastix} & \small{Group-RegNet} & \small{Proposed}\\
               
               \includegraphics[width=40mm,height=40mm]{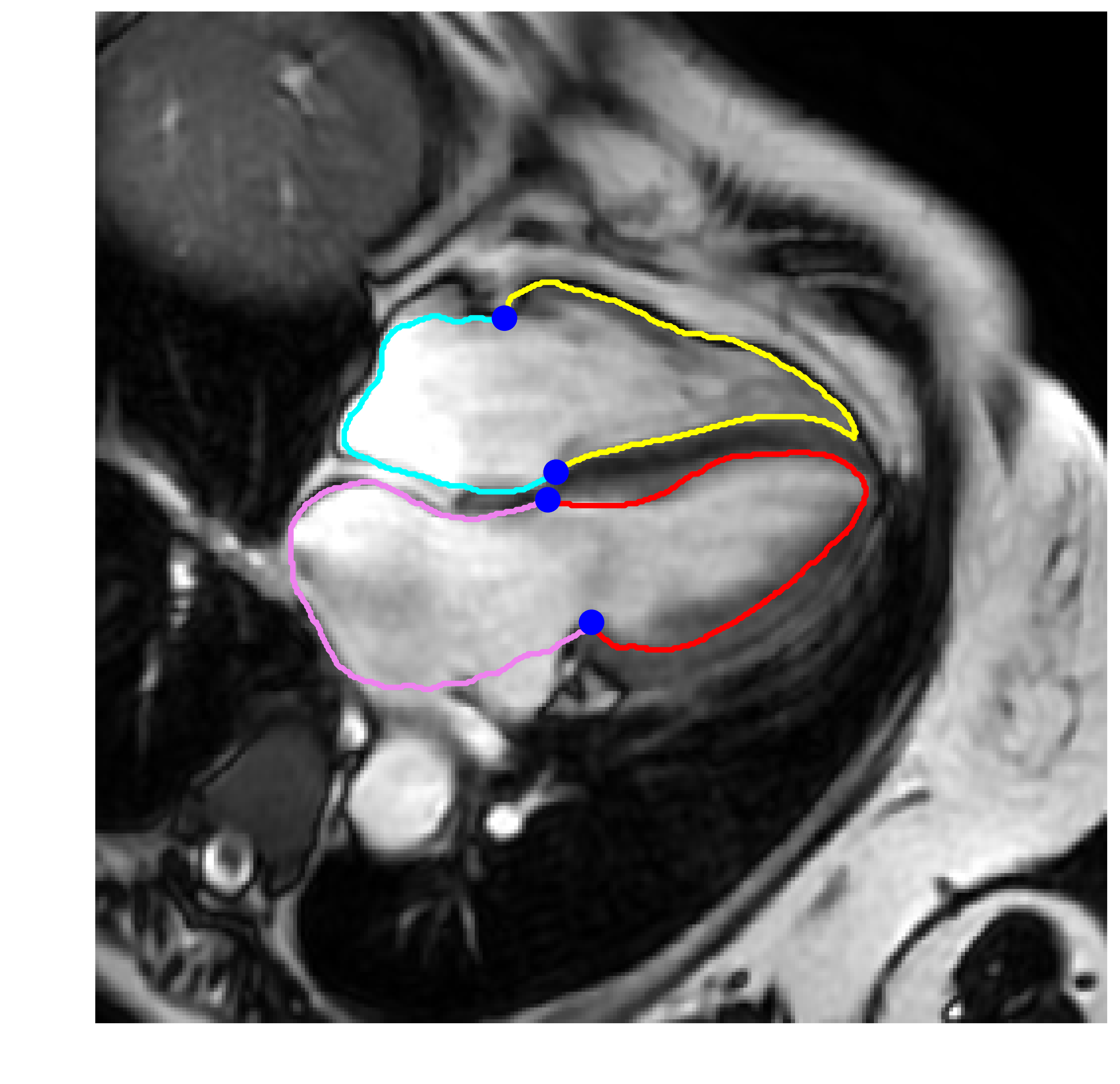} &
               \includegraphics[width=40mm,height=40mm]{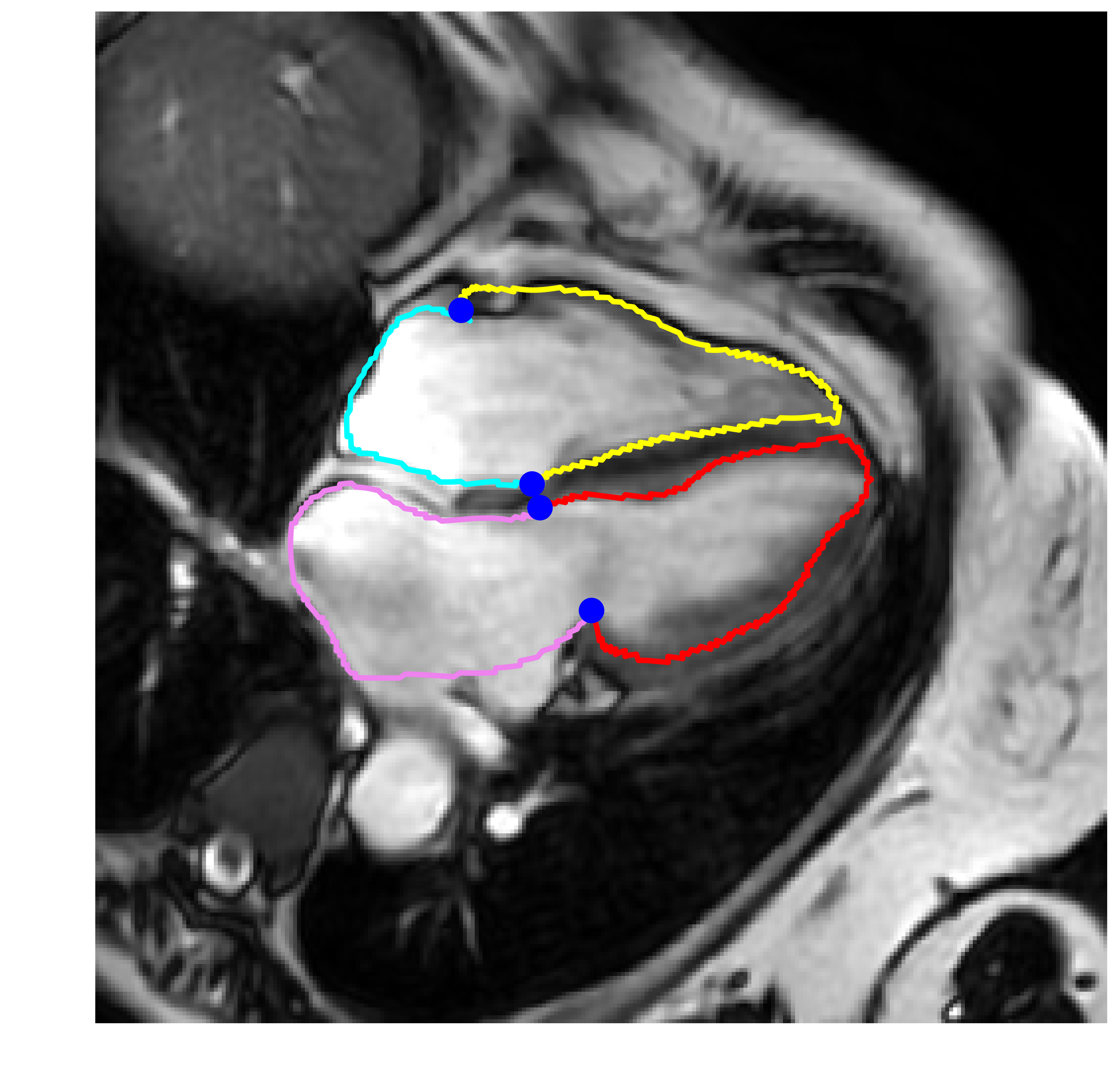}&
               \includegraphics[width=40mm,height=40mm]{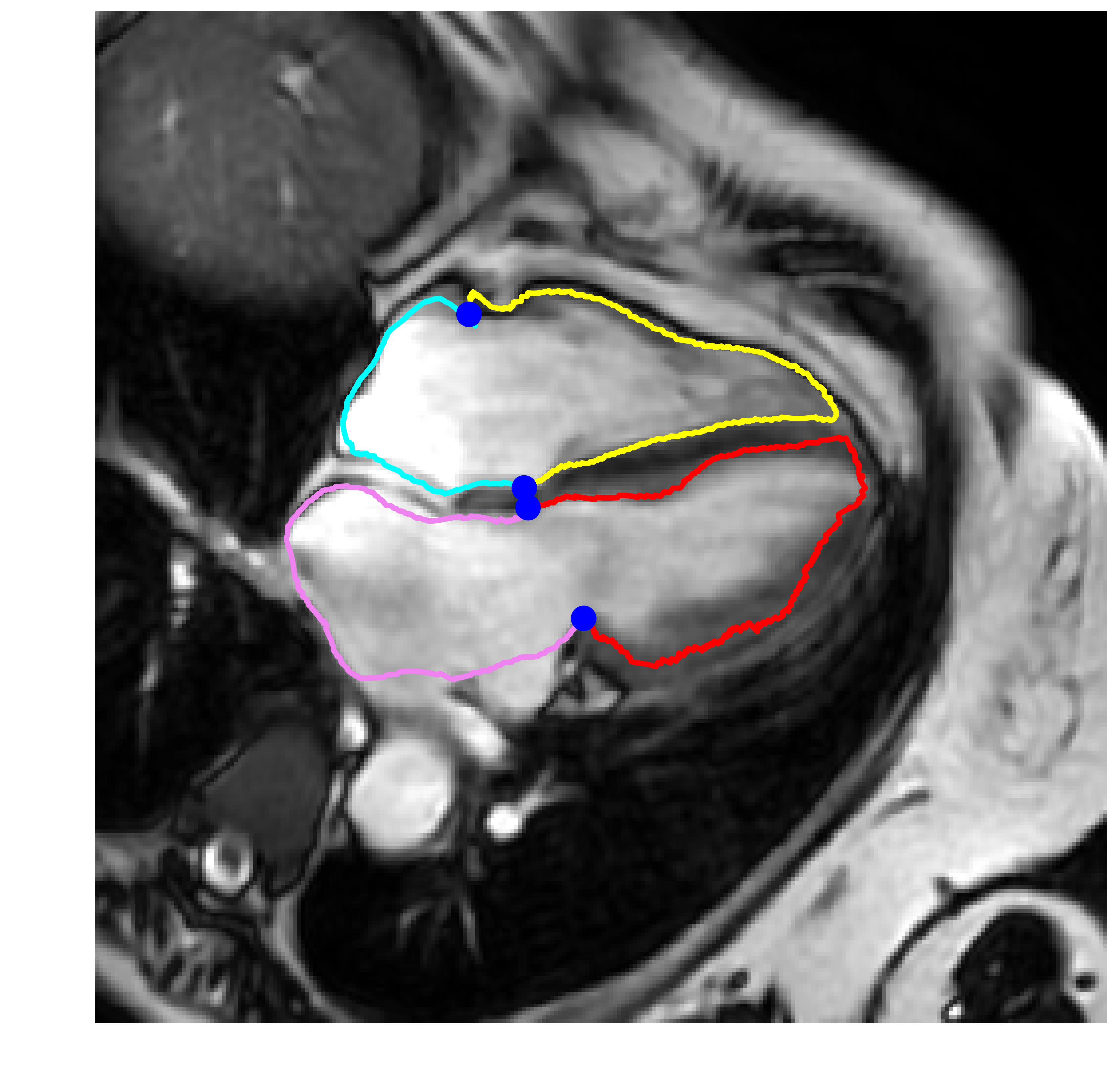}&
               \includegraphics[width=40mm,height=40mm]{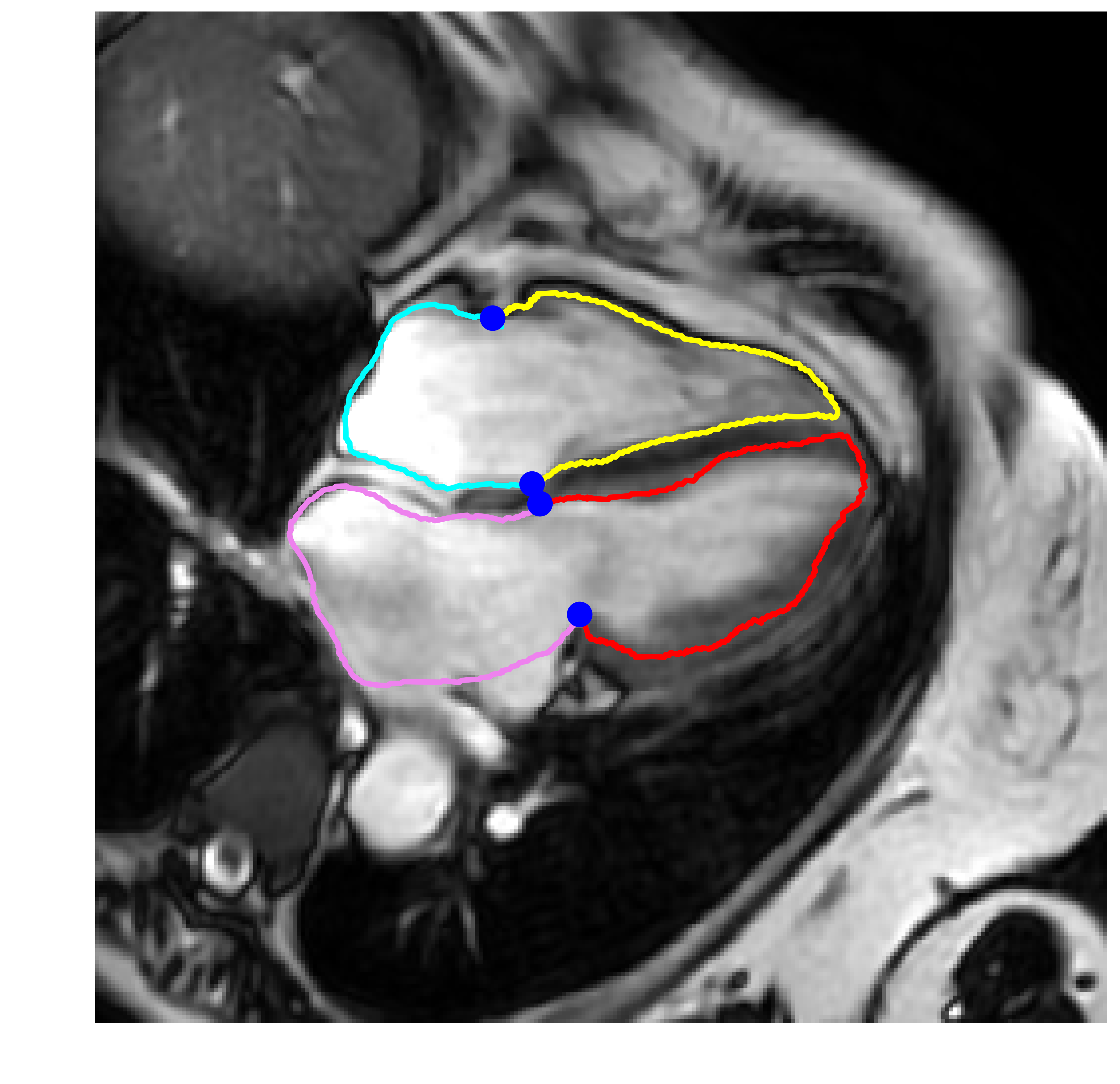} \\

               \includegraphics[width=40mm,height=40mm]{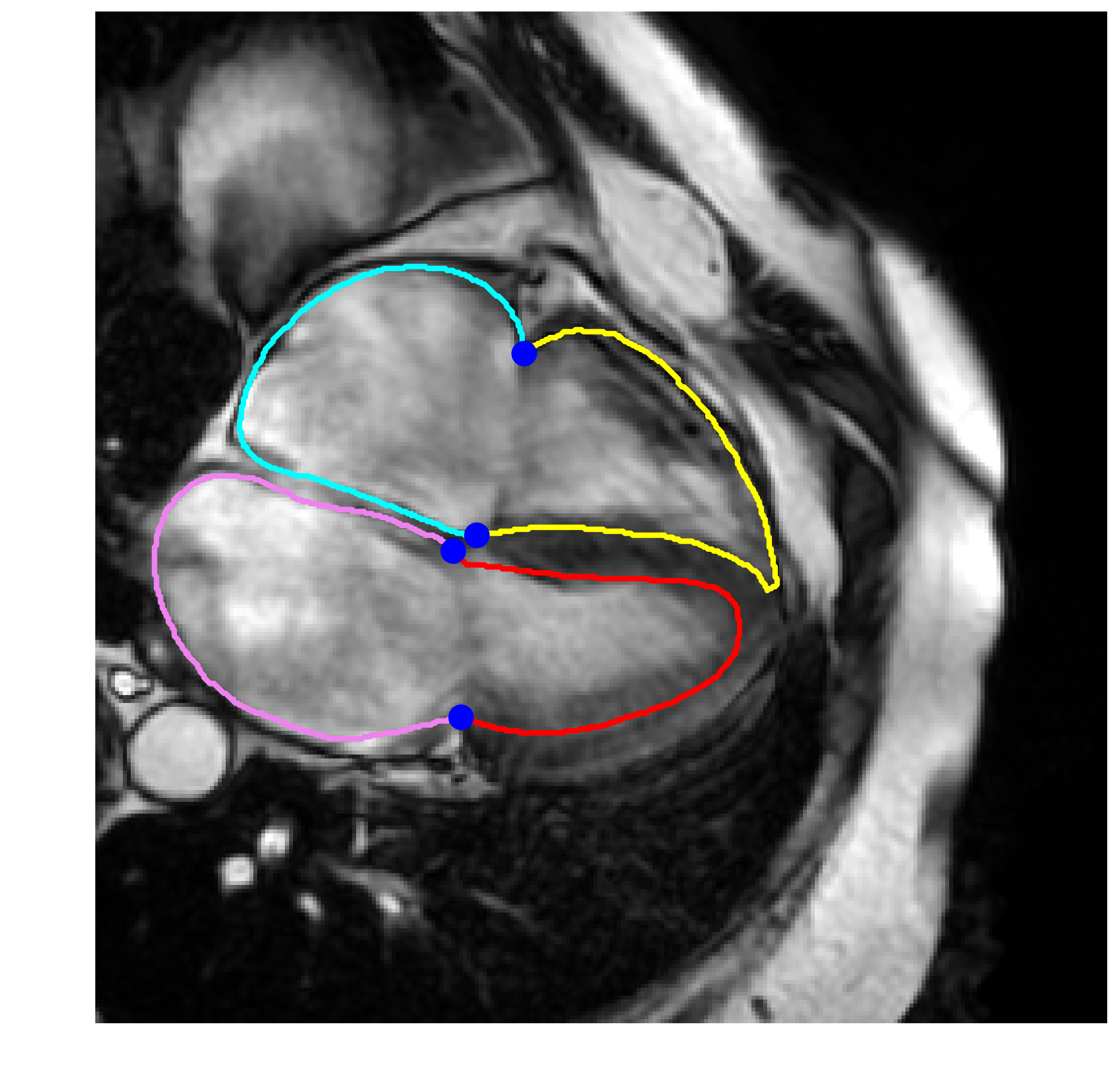} &
               \includegraphics[width=40mm,height=40mm]{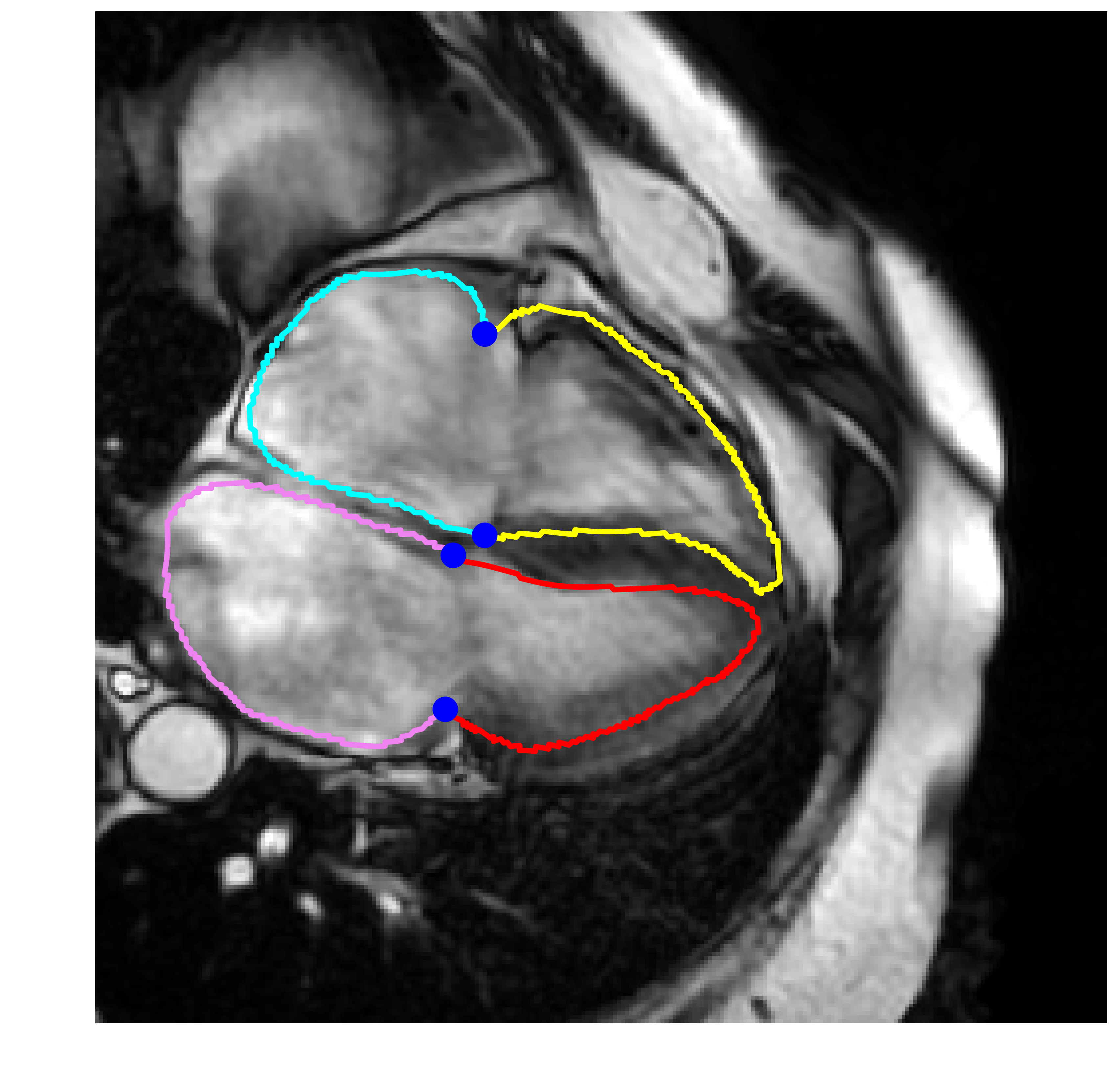}&
               \includegraphics[width=40mm,height=40mm]{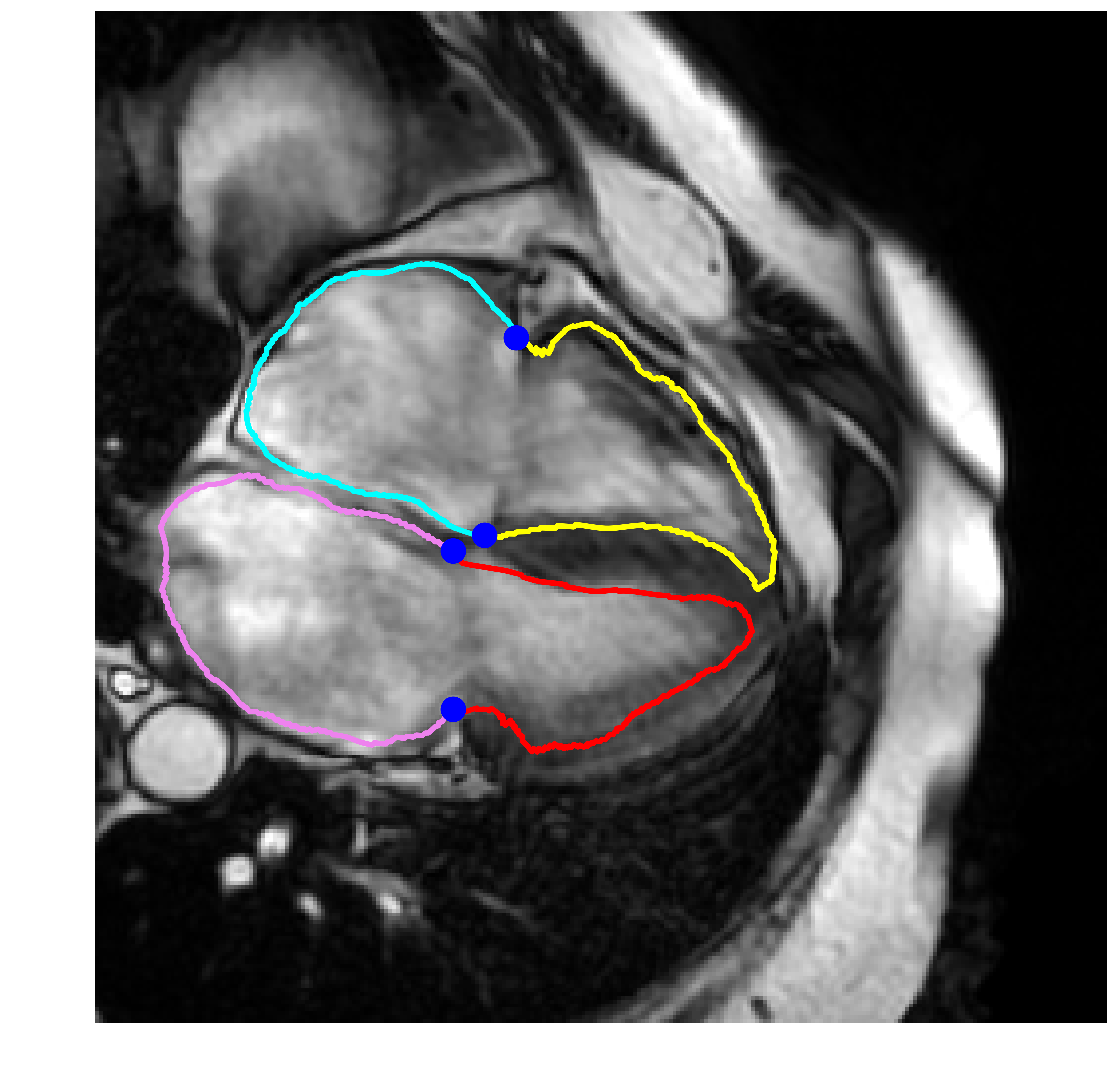}&
               \includegraphics[width=40mm,height=40mm]{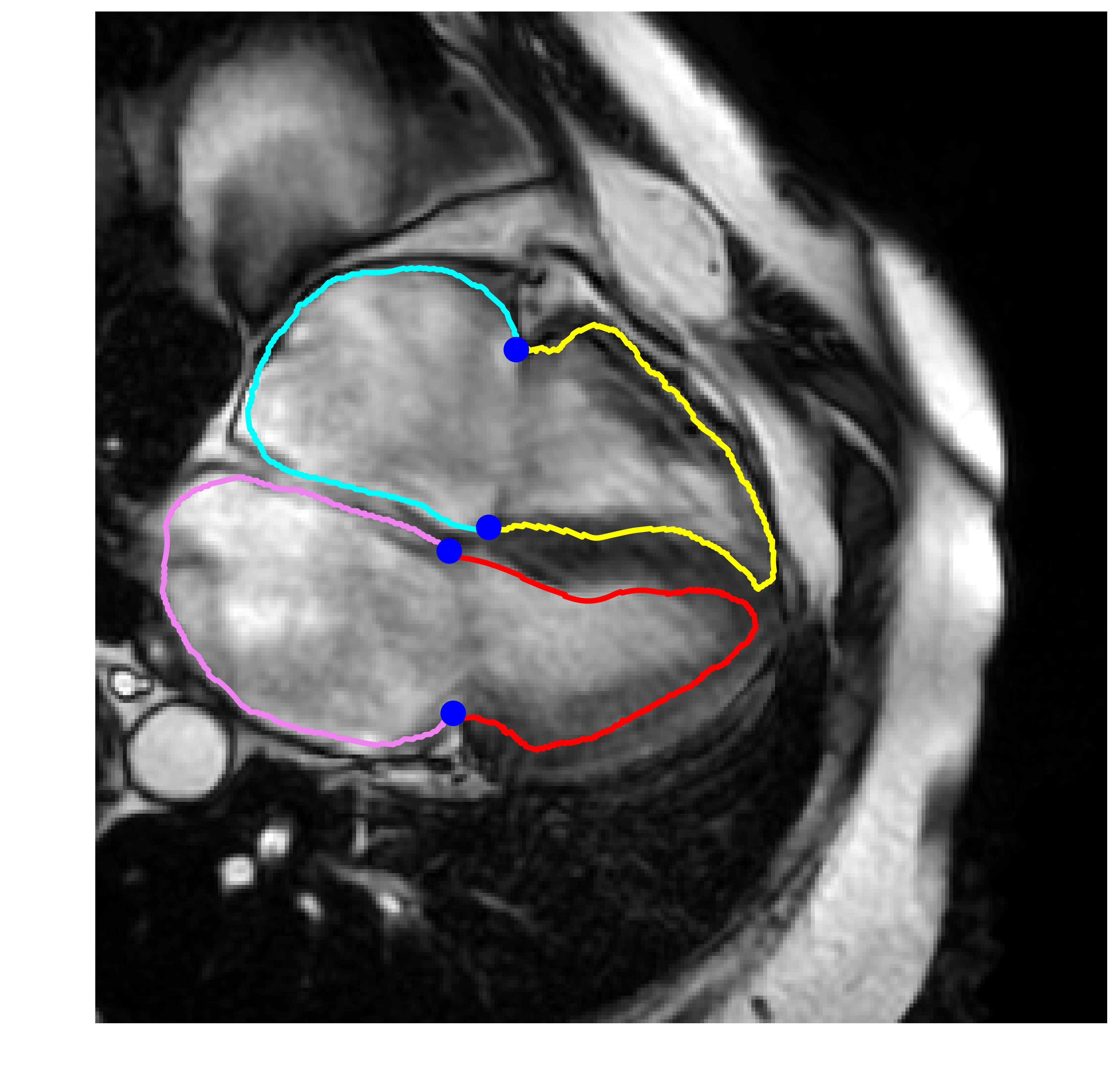} \\

               \includegraphics[width=40mm,height=40mm]{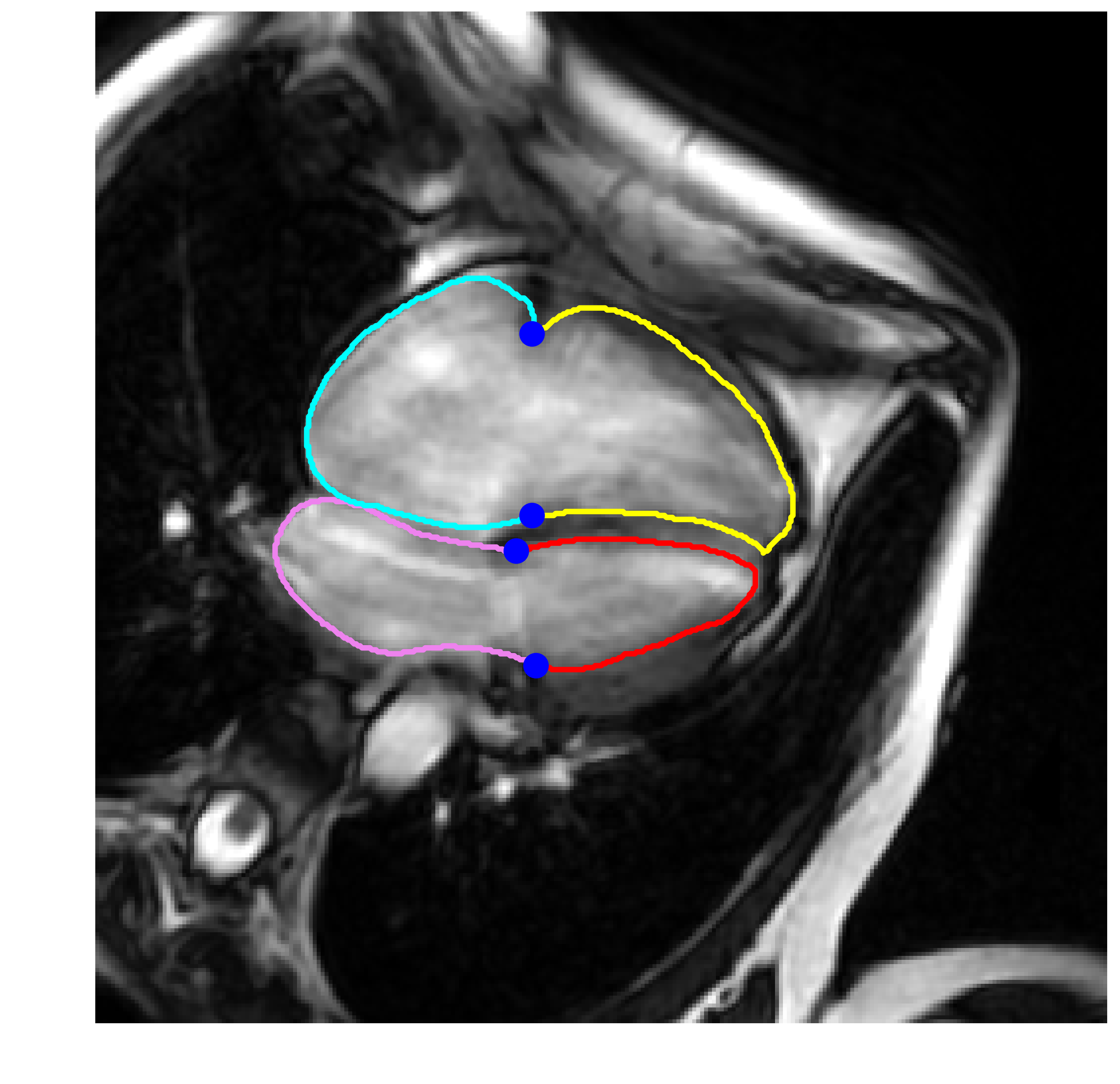} &
               \includegraphics[width=40mm,height=40mm]{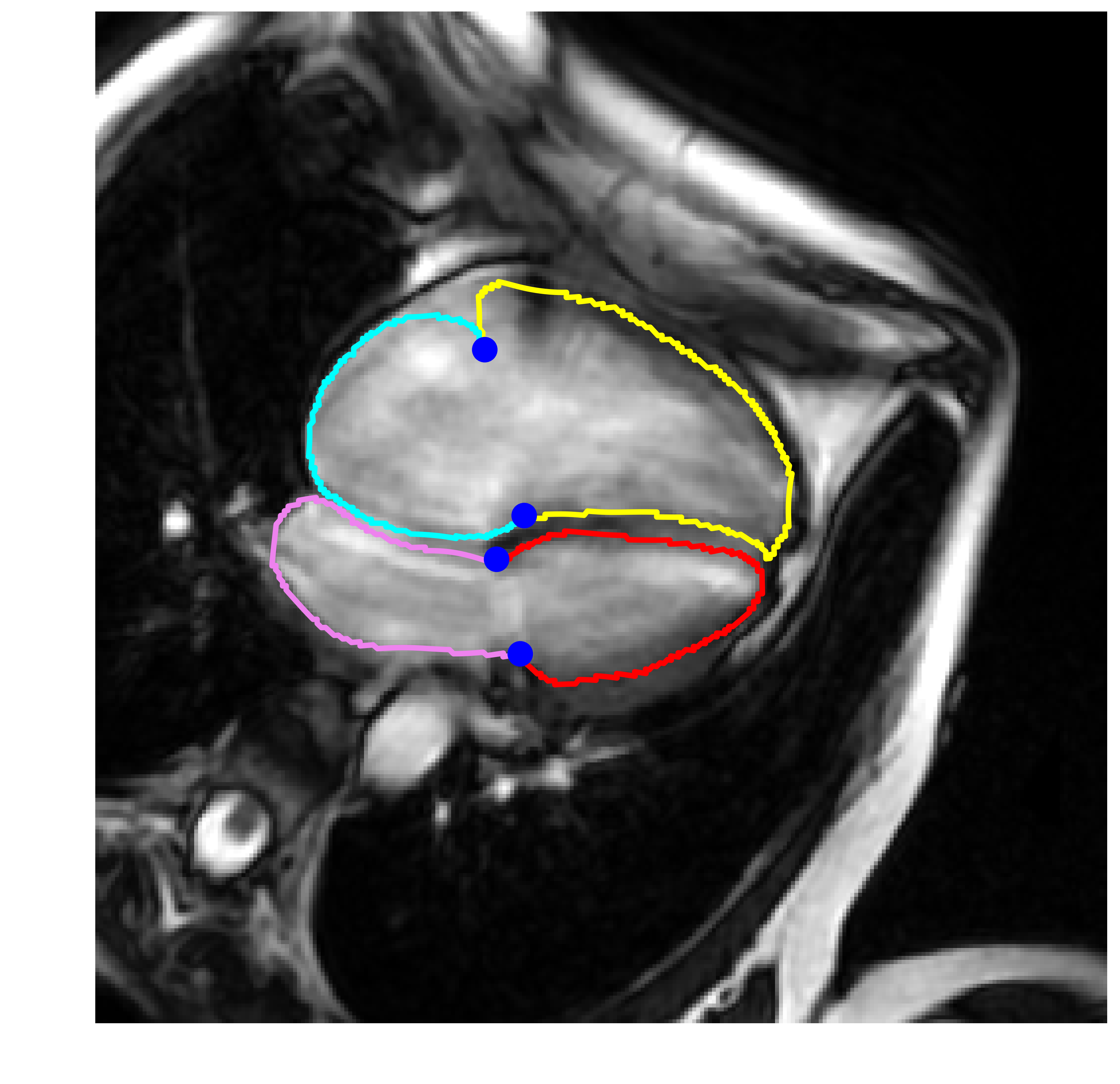}&
               \includegraphics[width=40mm,height=40mm]{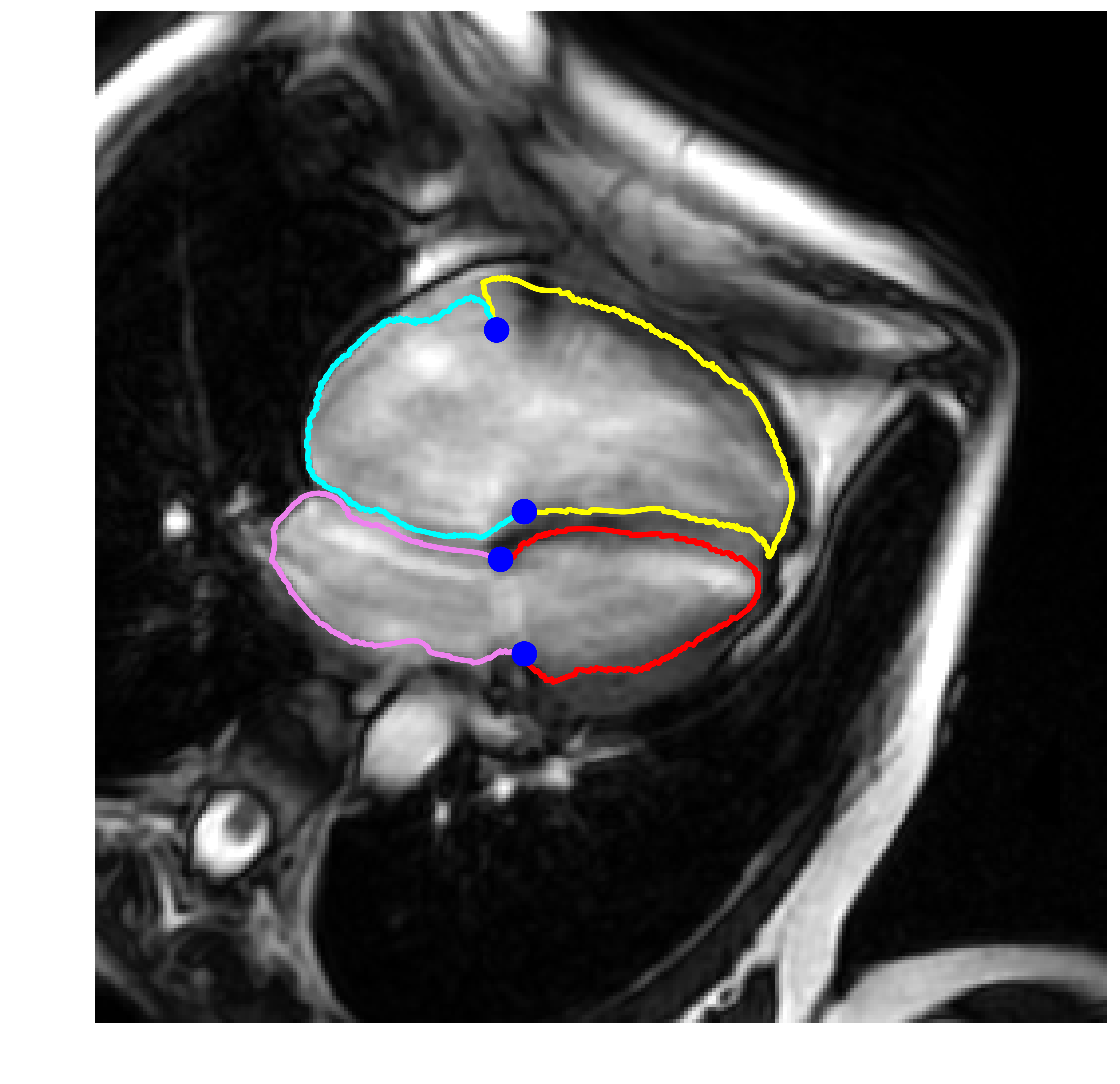}&
               \includegraphics[width=40mm,height=40mm]{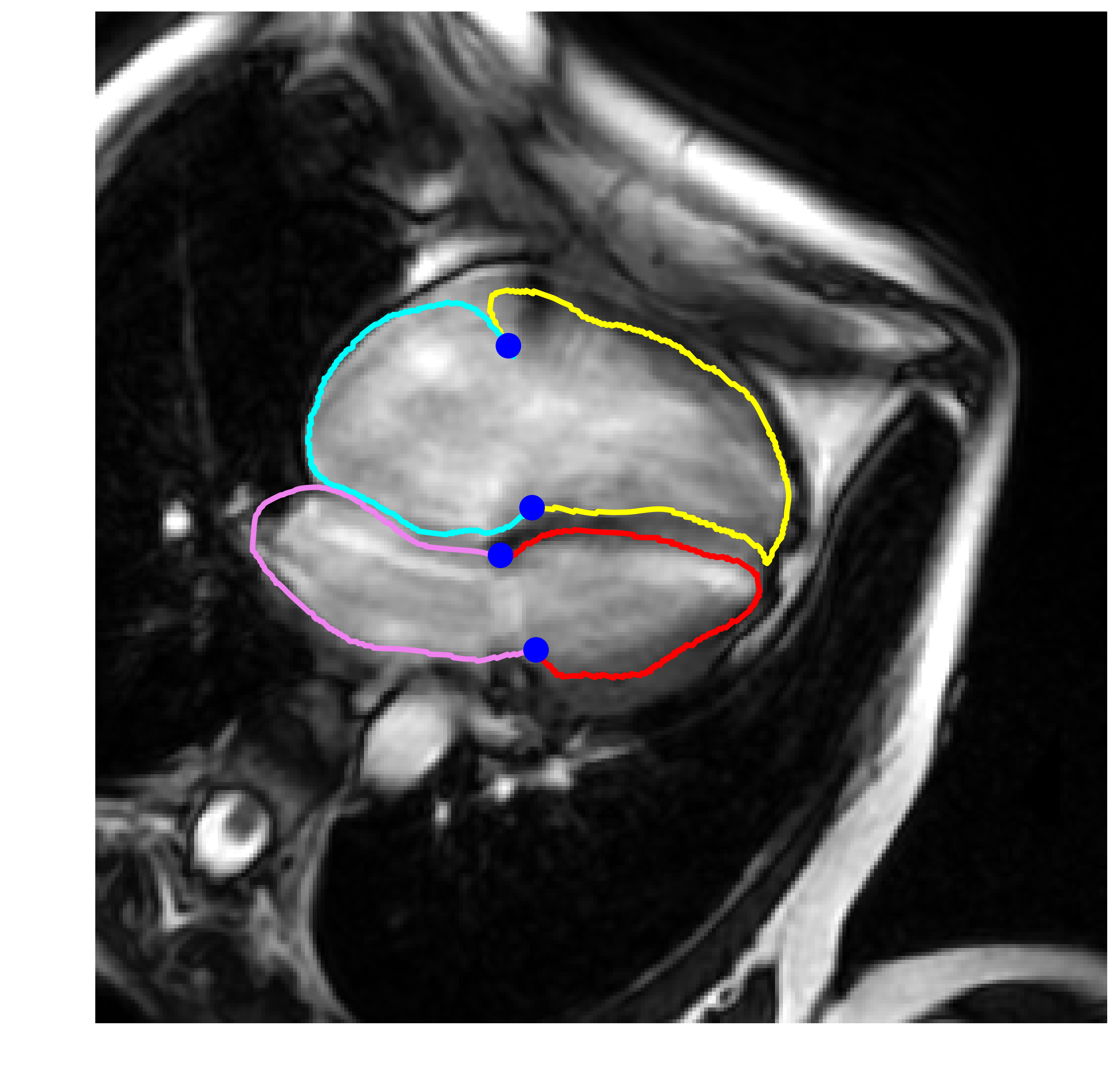} \\

		\end{tabular}
	}
\caption{Examples of the ES frame For different methods. From top to bottom are different subjects. From left to right, the manual contour, the propagated contour from the ED frame for different methods. The LV, RV, LA, and RA contours are colored in red, yellow, violet, and cyan, respectively. Additionally, the landmark points are marked in blue.}
\label{fig:comparison_examples}
\end{figure*}

\begin{figure*}[t]
\begin{center}
\includegraphics[width=\linewidth]{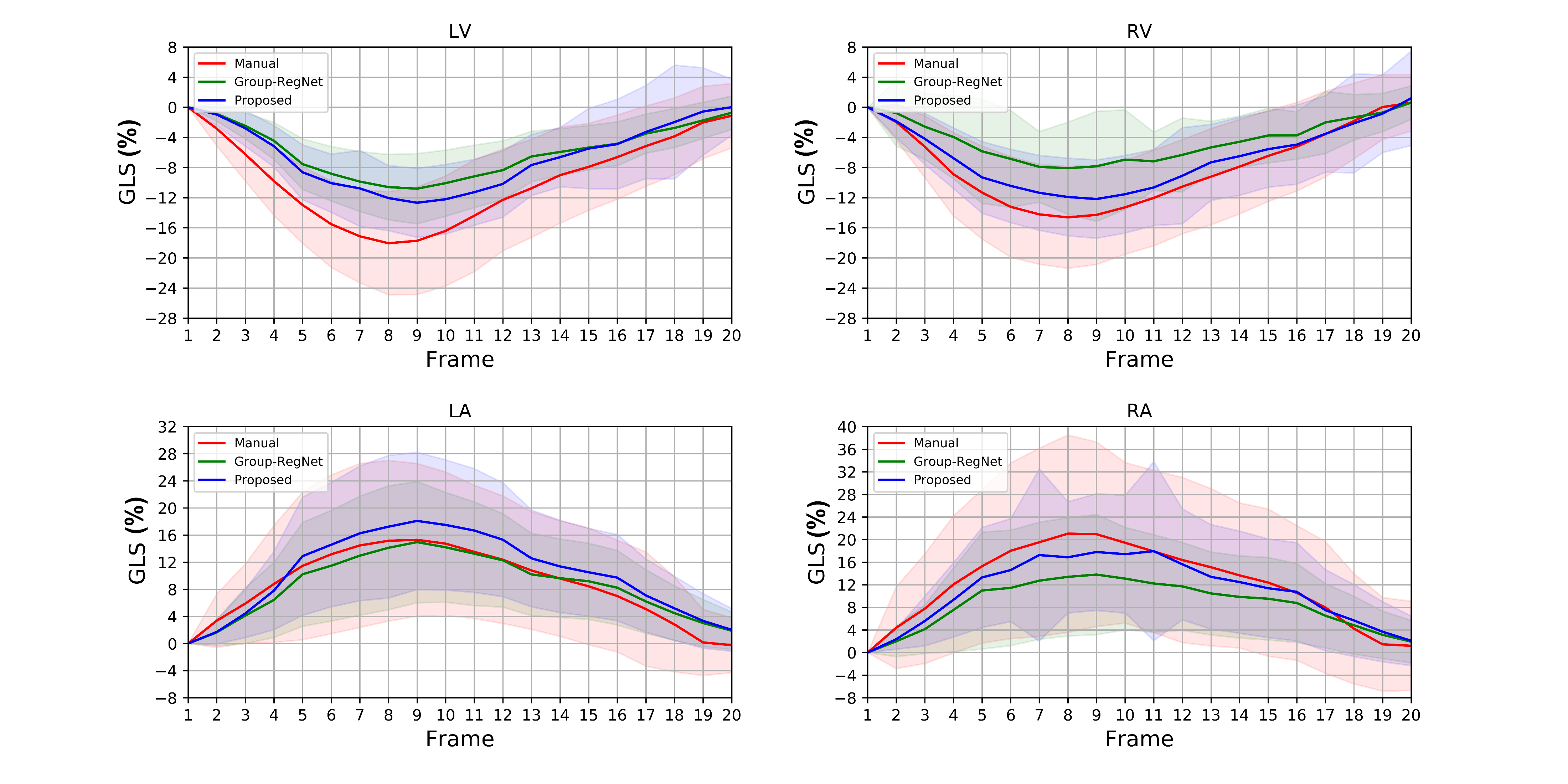}
\caption{Global longitudinal strain values for the validation dataset, where the solid line represent the mean across the patients and the shadow represent the standard deviation.}
\label{fig:strain}
\end{center}
\end{figure*}

\begin{figure*}[t]
\begin{center}
\includegraphics[width=\linewidth]{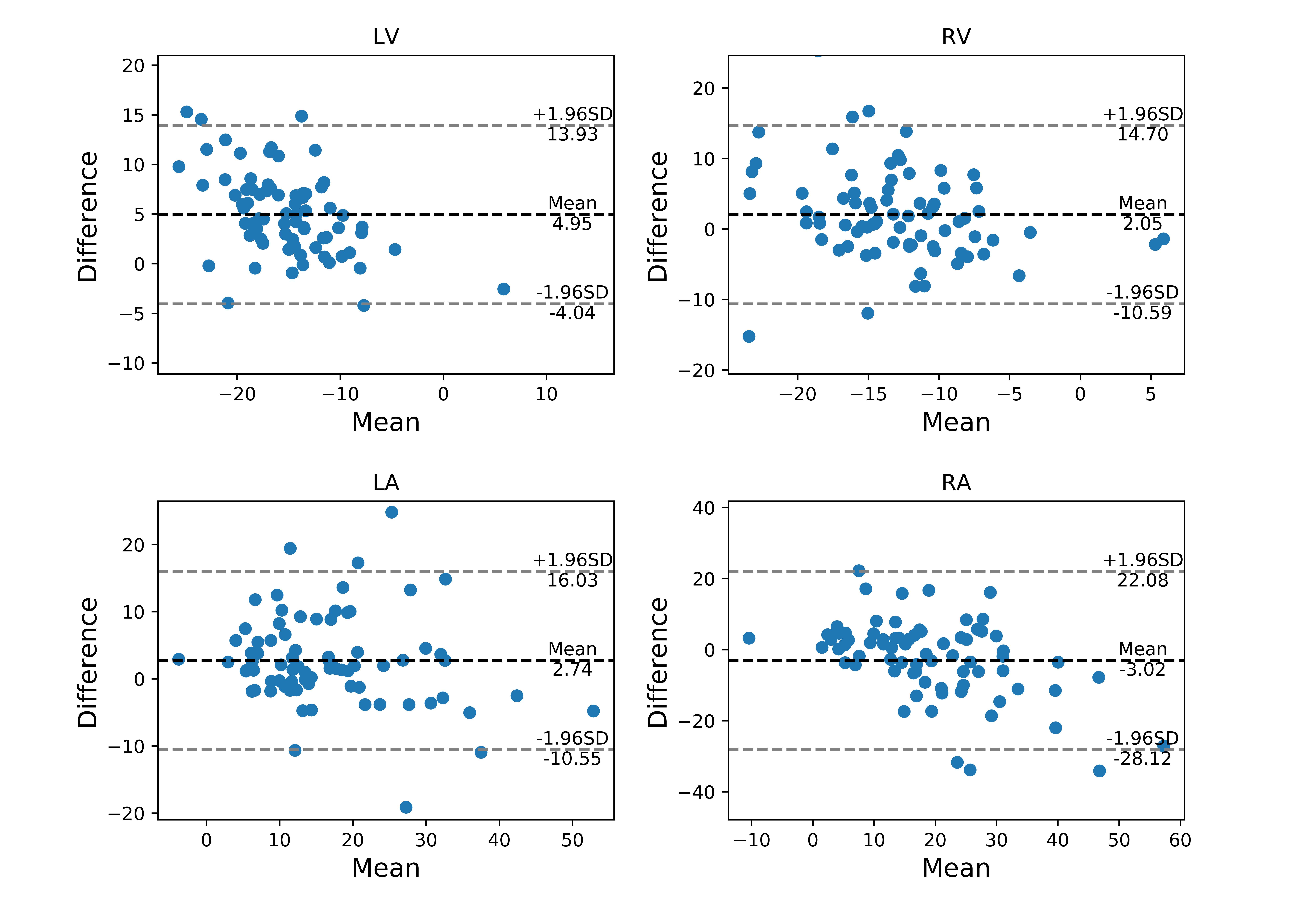}
\caption{Bland-Altman plot comparing GLS derived from proposed and manual contours at the ES phase.}
\label{fig:bland-altman}
\end{center}
\end{figure*}

\subsection{Optimizing CMRINet}
In this section, we examine different variations for CMRINet. For the first network, CMRINet$^a$, SegNet was frozen and the distance maps were derived from the manual contours. This network is therefore semi-automatic since it requires the availability of a manually annotated contour for one of the frames during inference time.

The second network, CMRINet$^b$, is a fully-automatic end-to-end network. SegNet is first trained so that it can estimate reasonably good contours, while Group-RegNet is frozen. Afterwards, Group-RegNet is trained, while SegNet is frozen. Finally, both SegNet and Group-RegNet are jointly trained. For this network, we propagated the automatically estimated contour from the first frame. This frame usually represents the ED phase, but not in all cases. 

The first two rows in Table \ref{table:joint_MCD} shows the MCD for CMRINet$^a$ and CMRINet$^b$. Both networks achieved comparable results, however CMRINet$^b$ is fully automatic and enables a fully automated quantitative analysis of cardiac function.  

The proposed CMRINet network utilizes SegNet for segmenting all available frames, enabling the propagation of any frame to the other frames (See Fig. \ref{fig:workflow}). The network, thereby generates a dictionary of contours that can be merged using ensemble methods such as majority voting. Additionally, the variance of the contours in the dictionary can potentially serve as an estimation of the model's uncertainty. The last row in Table \ref{table:joint_MCD} illustrates the mean contour distance (MCD) of the majority-voting ensemble of the dictionary, referred to as "Proposed". It achieves better results compared to CMRINet$^a$ and CMRINet$^b$. Therefore, this proposed method is used in the following comparisons.

\subsection{Comparison with Prior Work}
Our proposed network was compared to three existing methods from the literature. One of the methods is a conventional iterative method, while the other two are DL-based methods. It's worth noting that our proposed method is the only fully automated, end-to-end solution among them. For the iterative method, we compared our proposed network against a well-established open source image registration toolbox, named \texttt{elastix} \cite{elastix, metz2011nonrigid}. We used the NCC similarity loss with the settings proposed by \cite{shahzad2017fully}. 

For the DL-based methods, we chose two networks from \cite{elmahdy2021joint} namely the Dense and SEDD networks. Compared to the proposed CMRINet, Dense and SEDD segment and register the sequence in a parallel manner. This means that they are semi-automatic since they still require the availability of manually annotated contours for one of the frames. The Dense network models both segmentation and GW registration tasks with shared parameters except for task-specific ones in the output layer. The network is based on the same architecture as the base network, with different output layers. Dense parameter sharing prevents overfitting, but may not produce the best representation for individual tasks. The network has 1,895,592 trainable parameters. For the SEDD network, both segmentation and GW registration tasks share the same encoding path in the network, which then divides into two distinct decoder paths for each task. This specific network structure is referred to as the Shared Encoder Double Decoder (SEDD) network. The network is responsible for predicting both the displacement field from the registration path and the mask prediction from the segmentation path. SEDD has a total of 2,562,664 trainable parameters. For more details on training these networks and the losses used, see \cite{elmahdy2021joint}. 

Table \ref{table:comaprison_MCD} shows the comparison between the different networks in terms of MCD averaged across all frames as well as for the ES frames. For all the chambers, except for LV, the proposed network outperforms all the other networks including the joint SEDD and Dense networks. The MCD for the ES frame is slightly higher than the average, which is expected given that the propagated contour is the ED frame \review{(more details on the comparison in terms of DSC and HD metrics can be found in Appendix A).} Table \ref{table:landmark_MCD} shows the MCD for different landmark points. The proposed network is outperforming all other networks for the four landmark points. Same pattern occurs in the DSC and HD metrics as shown in Table \ref{table:comaprison_DSC} and \ref{table:comaprison_HD}.

Figure \ref{fig:proposed_examples} shows visual examples for the contours, DVF, and determinant of the Jacobian estimated from the proposed method. Figure \ref{fig:comparison_examples} shows a visual comparison between \texttt{elastix}, Group-RegNet and proposed network against the manually defined contours. It shows that the proposed method achieves better results compared to the other methods especially at the landmark points.

\subsection{Assessment of the DVF Quality}
The quality of the estimated DVF represented by the determinant of the Jacobian is important to ensure a smooth deformation field. Table \ref{table:dvf} shows the standard deviation of the determinant of the Jacobian and the folding fraction for different networks. The proposed network generated a better DVF with the lowest folding fraction. 

\subsection{Volume Analysis }
Table \ref{table:volume_table} shows the absolute difference between volumetric measures for LV such as ES (LVES) and ED (LVED) volumes and LVEF, estimated from different networks against the measures estimates from manual contours. The proposed network showed improvement against SegNet as well as other GW registration methods for all volumetric measures. 

\subsection{Strain Analysis}
Figure \ref{fig:strain} shows GLS strain values derived from manual contours against the propagated contours from both Group-RegNet and proposed networks. For the LV, both networks underestimated the GLS around the ES frames, with proposed slightly better than Group-RegNet. For the RV, GLS estimated from the proposed network was similar to manual results with a substantial improvement over Group-RegNet. For the LA, Group-RegNet showed a slight improvement over the proposed network compared to the GLS estimated from the manual contours. Figure \ref{fig:bland-altman} shows the Bland-Altman plots comparing GLS strain results derived from the proposed and manual contours. 

\begin{table}[t]
	\centering
	\caption[]{Comparison between the absolute difference between different networks and manual. Values in red represent the best results. \review{Daggers denote one-way ANOVA statistical significance for the proposed network against other networks.}}
	\resizebox{\linewidth}{!}{
 
        \begin{tabular}{lccc}
        \hline
         \multirow{2}{*}{Network} & \multicolumn{1}{c}{LVES volume (mL)} & \multicolumn{1}{c}{LVED volume (mL)} & \multicolumn{1}{c}{LVEF(\%)} \\ 
        &\multicolumn{1}{c}{$\mu \pm \sigma$} &\multicolumn{1}{c}{$\mu \pm \sigma$}&\multicolumn{1}{c}{$\mu \pm \sigma$}\\ \hline
        
        \texttt{elastix}& $12.2\pm 9.5$ & \textcolor{red}{$1.2 \pm 0.8$} & $4.7\pm 3.1$   \\ \hline
        
        Group-RegNet & $13.4\pm 10.6$ & $1.4 \pm 0.8$ & $4.6\pm 3.0$   \\ \hline

        Dense & $12.7\pm 9.6$ & $1.2\pm 0.9$ & $4.0\pm 3.2$   \\ \hline
        
        SEDD & $13.2\pm 10.3$ & $1.3\pm 0.6$ & $4.7\pm 3.3$   \\ \hline

        SegNet & $11.5\pm 9.4$ & $2.8\pm 2.3$ & $4.2\pm 3.4$   \\ \hline
        
        Proposed& \textcolor{red}{$11.0\pm 8.4^{\dagger}$} & $2.1\pm 1.5$ & \textcolor{red}{$3.5\pm 2.7^{\dagger}$}   \\ \hline
        
        \end{tabular}
        }
\label{table:volume_table}
\end{table}

\section{Discussion}\label{discussion_section}
In this study, we propose CMRINet to jointly optimize segmentation and groupwise registration tasks in an end-to-end network. The network enables fully automated cardiac function quantification in terms of volume and strain. By jointly optimizing segmentation and registration tasks we can capitalize on their respective strengths and minimize their limitations through directing the groupwise registration network with the aid of the underlying anatomical structures and mechanics. The proposed architecture comprises of two networks, namely SegNet and Group-RegNet. We optimized the segmentation network (SegNet) in terms of the training scheme and loss function. Additionally, we evaluated different template estimation methods for the Group-RegNet network. We compared our proposed network to iterative-based as well as DL-based methods. Finally, we evaluated our network in terms of clinically relevant volumetric parameters and global longitudinal strain.

We have trained and validated various networks on a large cohort of 4-chamber LAX view cine-MRI scans. The dataset is versatile and comprises a wide range of cardiac pathologies. The scans are from two MRI scanner vendors and exhibit challenging diversity in terms of orientations, heart size, contrast, noise, and field of view. 

For the segmentation network, SegNet, we defined a 2.5D training scheme to train the network in order to promote continuity in the contour prediction. The input to the network consisted of one, three or five frames, while the network only segmented the center frame. Our proposed 2.5D training scheme outperformed the 2D counterpart by incorporating information from adjacent frames. This enabled the network to learn the anatomical correlation between the frames and project it onto the segmentation prediction. However, including more neighboring frames deteriorated the network performance. We hypothesized that adding more frames forces the network to encode more information from distant anatomical frames with respect to the cardiac cycle. We also experimented with different loss functions and found that the sum of Dice similarity coefficient (DSC) and cross-entropy (CE) loss leveraged the foreground-background class imbalance while controlling the trade-off between false positives and false negatives. These results are consistent with the literature \cite{taghanaki2019combo}. 

For the groupwise registration network, Group-RegNet, we experimented with two template estimation methods namely average and PCA. In terms of MCD, the average template achieved better results. This might be due to the fact that cine-MRI registration is a mono-modal problem, therefore estimating the implicit template using the average of the warped images would be more accurate than PCA estimation. On the other hand, the PCA method estimates the eigenvector associated with the largest eigenvalue and since the cine-MRI images have the same intensity distribution the resultant template would be very similar to the average template. 

The first and second rows of Tables \ref{table:comaprison_MCD} and \ref{table:landmark_MCD} highlight the fact that the performance of Group-RegNet is similar to that of the conventional iterative approach represented by \cite{metz2011nonrigid} and implemented in \texttt{elastix}. In contrast to \texttt{elastix}, Group-RegNet is a learnable network that can encode and learn from the underlying data distribution and potentially generalize to unseen data. Additionally, Group-RegNet's runtime is $\sim$0.2 seconds compared to $\sim$20 seconds for \texttt{elastix}. 

For the joint segmentation and groupwise registration networks, we proposed to jointly model the segmentation and groupwise registration network in a single end-to-end network dubbed CMRINet. The proposed network was able to generate the segmentation and DVF in a fully automatic manner without the need of any manual annotations. Table \ref{table:joint_MCD} reveals that guiding CMRINet with the distance maps from the manual contours results in similar performance as compared to using the predicted contours from SegNet. This indicates that encoding contours in the form of distance maps can tolerate small error in segmentation prediction without affecting the overall performance of the network. Moreover, we proposed to propagate all predicted segmentations from SegNet to all other time frames, leveraging the predicted DVF. This proposed method resulted in a dictionary of contours for each frame in the cine-MRI sequence. By combining these contours via majority voting, we were able to further improve the results.  

We compared our proposed method to two joint segmentation and registration networks from \cite{elmahdy2021joint}. We found that guiding the GW registration through a dense network architecture achieved better performance than with a double decoder (SEDD) network, particularly for LA and RA. This indicates that sharing knowledge across all layers of the network is more beneficial than sharing it only in the encoding part. Furthermore, it suggests that both segmentation and GW registration tasks are highly correlated and learning them together can improve performance and regularize the network. In contrast to our proposed method, both the Dense and the SEDD networks required the availability of a manual contour for one frame. Tables \ref{table:comaprison_MCD} and \ref{table:landmark_MCD} shows that all methods performed similarly for LV contour detection, while the proposed method \review{resulted in a statistically significant improvement compared to }the other methods for LVM, RV, LA, RA as well as for all the landmark points. This indicates that encoding anatomical information in terms of distance maps is more efficient than encoding it through binary masks. Additionally, we checked the cases with high error in terms of MCD, and the common factor between them was the presence of high deformation especially in the LV chamber compared to the rest of the dataset. In terms of the scanner effect, it's important to note that although the number of cases acquired with the Siemens scanner is significantly lower in the training dataset compared to GE, this did not affect the network's performance on the Siemens cases in the validation set. This is a promising indication that the network may be able to generalize to data from other scanners as well.

% Moreover, we proposed to incorporating the manual contours in the GW registration network through distance maps, resulting in superior performance for all structures compared to using only masks. All the previous networks required the availability of the manual contour of at least one frame (in this paper it is ED frame), which is a tedious and time consuming task and prone to human error. For our last network, we proposed to automatically segment all the frames and thereby making the method fully automatic. However, the end-to-end network showed a small deterioration in performance compared to the semi-automatic network, possibly due to suboptimal predicted segmentation quality due to network errors, which propagate into the GW registration network. Nonetheless, the end-to-end approach enabled us to create a contour dictionary. The majority vote ensemble of these contours achieves performance similar to that of the proposed network but is more robust to errors propagated from the segmentation network.

Regarding the smoothness of the estimated DVF, as presented in Table \ref{table:dvf}, the proposed network \review{achieved a statistically significantly better performance compared to the other networks in terms of the standard deviation of the Jacobian as well as for the folding fraction.} We believe that incorporating the anatomical information from the segmentation network into the GW registration network served as an additional regularization method for the GW registration network. This resulted in higher quality DVF, which in turn yields better propagated contours and ultimately lower error.

% We evaluated the proposed network in terms of volumetric parameters such as LVES, LVED, and LVEF. For LVES, the GW registration networks achieved similar performance with a high bias since ES frame has the highest deformation from the propagated frame. On the other hand, SegNet and proposed method performed slightly better. This is potentially due to the anatomical guidance during training. \texttt{elastix}, Group-RegNet, Dense, SEDD networks propagates the first frame in the sequences, which often is the ED frame, therefore they achieve better perormance than SegNet and the proposed network.  Regarding LVEF, the proposed network achieved the best performance, this due to its consistency prediction between ES and ED volume prediction.
We evaluated the proposed network using volumetric parameters such as LVES volume, LVED volume, and LVEF. In terms of LVES, the GW registration networks demonstrated similar performance with a relatively high bias, because the ES frame has the highest deformation compared to the propagated frame. SegNet and our proposed method performed slightly better, which may be due to the anatomical guidance during training. The registration method using \texttt{elastix} and the networks Group-RegNet, Dense, and SEDD propagate the first frame in the sequence, which is often the ED frame, resulting in better performance than SegNet and our proposed network for LVED assessment. Regarding LVEF, our proposed network achieved the best performance due to its consistent prediction in ES and ED volume.

% The end-to-end network showed a small deterioration in performance compared to the semi-automatic network. This might be due to the fact that the quality of the predicted segmentation is not as optimal as the manual contour and prone to network's error, which then fed and propagated into the GW registration network. Additionally, the end-to-end approach enabled us to build a dictionary of contours. The majority vote ensemble of these contours is similar to the proposed network performance, however, it is more robust to the error propagated from the segmentation network.  

We analyzed the global longitudinal strain of different heart chambers, as illustrated in Fig. \ref{fig:strain}. The evaluation of strain across all chambers based on manual contours revealed a high variance across patients, indicating the significant diversity and challenge presented by the validation dataset. Group-RegNet and the proposed network achieved similar performance for the LV, which is consistent with the MCD values presented in Tables \ref{table:comaprison_MCD} and \ref{table:landmark_MCD}. However, both networks underestimated the strain, particularly during the ES phase. This could be due to the smoothness of the estimated DVF, which limited the network's ability to learn very high deformations around the ES phase. For the RV, the proposed network yielded results that were very close to the manual GLS, while Group-RegNet revealed an underestimation of GLS, which is again consistent with the MCD values. Regarding the LA, both the proposed network and Group-RegNet showed very similar results to the manual GLS. For RA, The proposed network had similar distribution to the manual GLS, which is consistent with the reported MCD errors. 

\review{In this study, while our proposed joint segmentation and registration method demonstrated considerable advancements in performance, it is important to note several limitations that shaped the scope of our evaluation. Firstly, the segmentation results of other state-of-the-art segmentation-only methods, such as nnU-Net \mbox{\cite{isensee2021nnu}} or more advanced network structures, and the registration outcomes of dedicated registration methods like VoxelMorph \mbox{\cite{balakrishnan2019voxelmorph}} or other advanced techniques, were not directly included in our evaluation. The purposeful exclusion of such comparisons stemmed from our primary focus on showcasing the collective improvement achieved by our method in contrast to dedicated segmentation networks. While acknowledging the potential for enhanced results by utilizing more advanced networks in individual segmentation or registration components, this comparative analysis was not the primary goal of our study.

Moreover, a crucial limitation stems from the constrained size of the annotated dataset, which restricted the creation of an independent test dataset for comprehensive validation. The limited availability of annotated data constrained the extent to which we could generalize the performance of our method across a larger and more diverse dataset. However, it's important to highlight that efforts are underway to address this limitation by actively acquiring more annotated data. This ongoing process aims to improve the robustness and generalizability of our proposed method by incorporating a more extensive and varied dataset for future evaluations.

Future research endeavors could leverage larger annotated datasets to conduct a broader comparison across various segmentation-only and registration-only methods, enabling a more comprehensive understanding of performance disparities and further validating the efficacy of our joint approach. Additionally, a promising direction for future exploration involves optimizing the network architecture based on the final strain performance. Moreover, further investigations could delve into the effective combination of various contours from the contour dictionary to attain optimal performance. These future endeavors hold potential to enhance the robustness and applicability of our method within the domain.}

\section{Conclusion} \label{conclusion_section}
In this paper, we introduce an end-to-end joint segmentation and GW registration network that performs fully automated quantification of cardiac function from cine-MRI. 
% Our network involves training both segmentation and GW registration networks simultaneously. We have developed and optimized training scheme and loss function for the segmentation network. We also experimented with two different template estimation methods for the GW registration network and compared our proposed network with different methods from literature.
We demonstrate that training segmentation and GW registration networks jointly produces superior results compared to training these networks individually. Our proposed CMRINet network can independently predict masks for all temporal frames and the DVF. This facilitates the propagation of all frames, enabling users to choose from a dictionary of N masks and also perform strain quantification. The network achieved an average contour distance of 1.7 mm, 1.0 mm, 1.1 mm, 1.1 mm, and 1.1 mm for LV, LVM, RV, LA, and RA, respectively.

The fast and simultaneous prediction of cardiac contours as well as strain by our proposed network allows for a full cardiac assessment in one pass and subsequently  streamline the quantitative assessment of cardiac function in the clinic.

% In this paper, we propose an end-to-end anatomically-guided GW registration network, where a segmentation and GW registration networks are trained jointly. We optimized the training scheme as well as loss function for the segmentation network. Additionally, we experimented with two template estimation approaches for the GW registration network. Moreover, we compared different anatomically-guided GW registration architectures. 

% We showed that training segmentation and GW registration networks simultaneously outperforms both segmentation and GW registration networks individually. The proposed fully autonomous anatomically-guided network was able to predict masks for for all the time point frames as well as the DVF, which eventually enabled the propagation of all frames to all frames. 

% The proposed network resulted in an average contour distance of 1.7 mm, 1.1 mm, 1.2 mm, 1.1 mm, 1.2 mm for LV, LVm, RV, LA, and RA, respectively. Since the network is able to predict the segmentation and DVF simultaneously, which gives the capabilities of doing a full cardiac assessment in a single go and potentially accelerate the clinical workflow. 
\section{Acknowledgment}
The authors would like to express their gratitude to Sheffield University Hospital, UK for providing the dataset and corresponding manual contours used in this study. This work was supported by an NIHR AI Award, AI\_AWARD01706.

%%Harvard
\bibliographystyle{apalike}
\bibliography{references}

\clearpage
\onecolumn
\appendix
\section{}
\vspace{0.5cm}
\label{appendix_section}
% In this appendix we provide a detailed results for the proposed method in terms of DSC and \%95 HD.
% \vspace{0.5cm}

\begin{table*}[h]
	\centering
	\caption[]{DSC values for \review{the registration output of different methods.} The Mean column  represents the average across all frames, while ES represents the average for the ES frames only. x$\pm$x represents mean$\pm$std. Values in red represent the best results. \review{Daggers denote one-way ANOVA statistical significance for the proposed network against other networks.}}
	\resizebox{\linewidth}{!}{
 
        \begin{tabular}{lcc|cc|cc|cc|cc}
        \hline
         \multirow{2}{*}{Network} & \multicolumn{2}{c}{LV} & \multicolumn{2}{c}{LVM} & \multicolumn{2}{c}{RV} & \multicolumn{2}{c}{LA} & \multicolumn{2}{c}{RA} \\ 

        & Mean & ES &  Mean & ES &  Mean & ES &  Mean & ES &  Mean & ES \\ \hline 
        
        \texttt{elastix}& $0.92\pm 0.03$ & $0.87\pm 0.06$ & $0.81\pm 0.05$ & $0.76\pm 0.08$ & $0.91\pm 0.03$ & $0.87\pm 0.06$ & $0.94\pm 0.03$ & $0.93\pm 0.04$ & $0.90\pm 0.06$ & $0.87\pm 0.08$ \\ \hline 

        Group-RegNet & \textcolor{red}{$0.92\pm 0.02$} & \textcolor{red}{$0.87\pm 0.05$} & $0.81\pm 0.05$ & $0.75\pm 0.08$ & $0.92\pm 0.03$ & $0.88\pm 0.05$ & $0.93\pm 0.03$ & $0.92\pm 0.04$ & $0.90\pm 0.07$ & $0.87\pm 0.09$ \\ \hline 

        Dense & $0.91\pm 0.03$ & $0.85\pm 0.05$ & $0.80\pm 0.05$ & $0.73\pm 0.08$ & $0.91\pm 0.03$ & $0.87\pm 0.06$ & $0.93\pm 0.03$ & $0.91\pm 0.05$ & $0.90\pm 0.08$ & $0.85\pm 0.12$ \\ \hline  

        SEDD & $0.92\pm 0.03$ & $0.87\pm 0.05$ & $0.81\pm 0.05$ & $0.76\pm 0.08$ & $0.92\pm 0.03$ & $0.88\pm 0.06$ & $0.93\pm 0.03$ & $0.92\pm 0.04$ & $0.90\pm 0.08$ & $0.86\pm 0.12$ \\ \hline 

        Proposed & $0.92\pm 0.02$ & $0.86\pm 0.05$ & \textcolor{red}{$0.82\pm 0.04$} & \textcolor{red}{$0.76\pm 0.07$} & \textcolor{red}{$0.93\pm 0.02$} & \textcolor{red}{$0.91\pm 0.04$} & \textcolor{red}{$0.95\pm 0.03^{\dagger}$} & \textcolor{red}{$0.94\pm 0.03^{\dagger}$} & \textcolor{red}{$0.92\pm 0.05^{\dagger}$} & \textcolor{red}{$0.90\pm 0.07^{\dagger}$}   \\ \hline 
        
        \end{tabular}
        }
\label{table:comaprison_DSC}
\end{table*}

\begin{table*}[h]
	\centering
	\caption[]{HD (mm) values for \review{the registration output of different methods}. The Mean column  represents the average across all frames, while ES represents the average for the ES frames only. x$\pm$x represents mean$\pm$std. Values in red represent the best results. \review{Daggers denote one-way ANOVA statistical significance for the proposed network against other networks.}}
	\resizebox{\linewidth}{!}{
 
        \begin{tabular}{lcc|cc|cc|cc|cc}
        \hline
         \multirow{2}{*}{Network} & \multicolumn{2}{c}{LV} & \multicolumn{2}{c}{LVM} & \multicolumn{2}{c}{RV} & \multicolumn{2}{c}{LA} & \multicolumn{2}{c}{RA} \\ 

        & Mean & ES &  Mean & ES &  Mean & ES &  Mean & ES &  Mean & ES \\ \hline 
        
        \texttt{elastix}& \textcolor{red}{$5.1\pm 1.6$} & \textcolor{red}{$7.2\pm 2.7$} & $5.1\pm 1.5$ & $7.0\pm 2.5$ & $6.3\pm 2.4$ & $8.8\pm 3.9$ & $4.1\pm 1.8$ & $5.3\pm 2.9$ & $6.4\pm 3.1$ & $9.3\pm 5.5$ \\ \hline 

        Group-RegNet & $5.4\pm 1.6$ & $7.8\pm 2.8$ & $5.3\pm 1.6$ & $7.4\pm 2.6$ & $6.2\pm 2.1$ & $8.3\pm 3.9$ & $4.5\pm 1.7$ & $5.8\pm 2.7$ & $6.3\pm 2.6$ & $8.8\pm 4.8$ \\ \hline 

        Dense & $5.7\pm 1.7$ & $8.5\pm 3.0$ & $5.4\pm 1.6$ & $7.8\pm 2.9$ & $6.7\pm 2.4$ & $9.5\pm 4.2$ & $4.7\pm 1.8$ & $6.3\pm 3.1$ & $6.9\pm 3.0$ & $10.3\pm 5.7$ \\ \hline  
 
        SEDD & $5.6\pm 1.6$ & $8.0\pm 2.8$ & $5.4\pm 1.6$ & $7.6\pm 2.7$ & $6.3\pm 2.3$ & $8.6\pm 3.9$ & $4.6\pm 1.7$ & $6.0\pm 2.8$ & $6.5\pm 2.9$ & $9.3\pm 5.3$ \\ \hline 

        Proposed & $5.7\pm 2.5$ & $8.2\pm 3.8$ & \textcolor{red}{$5.0\pm 2.1^{\dagger}$} & \textcolor{red}{$6.1\pm 3.1^{\dagger}$} & \textcolor{red}{$5.8\pm 2.7^{\dagger}$} & \textcolor{red}{$7.6\pm 4.8^{\dagger}$} & \textcolor{red}{$4.1\pm 1.8^{\dagger}$} & \textcolor{red}{$5.1\pm 2.8^{\dagger}$} & \textcolor{red}{$5.3\pm 2.4^{\dagger}$} & \textcolor{red}{$7.8\pm 4.0^{\dagger}$} \\ \hline 
        
        \end{tabular}
        }
\label{table:comaprison_HD}
\end{table*}

\end{document}